\pdfoutput=1

 \documentclass[10pt,twocolumn,twoside,journal]{IEEEtran}

\usepackage{cs}
\usepackage{mathsymb}
\usepackage{verbatim}
\usepackage{subfigure}
\usepackage{url}

\usepackage{graphicx}
\graphicspath{{./}{./Fig/}{./Figs/}{./Fig/eps/}{./Fig/pdf/}}

\def\bE{\operatorname{\mathbb E}}            

\usepackage{algorithm2e}
\let\savedalgorithm\algorithm
\let\savedendalgorithm\endalgorithm
\newenvironment{algorithmic}{%
\savedalgorithm
}{%
\savedendalgorithm
}

\usepackage{algorithm}

\usepackage{amsmath}
\usepackage{url}
\usepackage{multirow}

\def\Integer{\mathbb{Z}^+}

\begin{document}

\title{Incremental Training of 
       a Detector Using Online Sparse Eigen-decomposition}

\author{
         Sakrapee Paisitkriangkrai,
         Chunhua~Shen,
         and
         Jian Zhang,~\IEEEmembership{Senior Member,~IEEE}
\thanks
{ 
Manuscript received April X, 200X; revised March X, 200X.
NICTA is funded through the Australian Government's 
{\em Backing Australia's Ability} initiative, 
in part through the Australian Research Council.           
The associate editor coordinating the review of this manuscript
and approving it for publication was Dr. X X. 
}
\thanks
{
S. Paisitkriangkrai
and J. Zhang are with NICTA, Neville Roach Laboratory,
Kensington, NSW 2052, Australia, and also with the University of New
South Wales, Sydney, NSW 2052, Australia
(e-mail: \{paul.pais, jian.zhang\}@nicta.com.au). 
}
\thanks
{
C. Shen is with NICTA, Canberra Research Laboratory, 
 Canberra, ACT 2601, Australia,
and also with the Australian National University, Canberra,
ACT 0200,Australia
(e-mail: chunhua.shen@nicta.com.au).
}
\thanks
 {
 Color versions of one or more of the figures in this paper are available online
 at http://ieeexplore.ieee.org.
 }
} 

\markboth{IEEE TRANSACTIONS ON IMAGE PROCESSING,~VOL.~XX, NO.~XX,~MARCH~200X}
{Paisitkriangkrai
\MakeLowercase{\textit{et al.}}: Incremental Training a Detector Using Online Sparse Eigen-decomposition}

\maketitle

\begin{abstract}

    The ability to efficiently and accurately detect objects plays a very crucial role
    for many computer vision tasks. 
    Recently, offline object detectors have shown a tremendous success.
    However, one major drawback of offline techniques is that a complete set of training data has to
    be collected beforehand. In addition, once learned, an offline detector can not make use of
    newly arriving data. To alleviate these drawbacks, online learning has been adopted with the
    following objectives: (1) the technique should be computationally and storage efficient; 
    (2) the
    updated classifier must maintain its high classification accuracy. 
    In this paper, we propose an
    effective and efficient framework for learning an adaptive 
    online greedy sparse linear discriminant
    analysis (GSLDA) model. Unlike many existing online boosting detectors, 
    which usually apply exponential or logistic loss,
    our online algorithm makes use of LDA's learning criterion that not only aims to maximize the
    class-separation criterion but also incorporates the asymmetrical property of training data
    distributions. 
    We provide a better alternative for online boosting algorithms in the context of training a 
    visual object detector. 
    We demonstrate the robustness and efficiency of our methods on handwriting digit and face
    data sets. Our results confirm that object detection tasks 
    benefit significantly when trained in
    an online manner.

\end{abstract}


\section{Introduction}

  \IEEEPARstart{R}{eal-time} object detection plays an important role in many real-world vision
  applications. It is used as a preceding step in applications such as intelligent video
  surveillance, content based image retrieval, face and activity recognition. 
  Object detection is a challenging problem due
  to the large variations in visual appearances, object poses, illumination, camera motion, \etc.
  All these issues have made the problem very challenging from a machine vision perspective.

  The literature on object detection is abundant. A through discussion on this topic can be found in
  several surveys: faces \cite{Yang2002Detecting,Hjelmas2001Face}, human and pedestrians
  \cite{Munder2006Pedestrian,Geronimo2009Survey}, eyes \cite{Campadelli2007Eye}, vehicles
  \cite{Sun2006Onroad}, \etc. In this paper, we review only the most relevant visual detection work,
  focusing on algorithms that operate directly on classification based visual object detection and
  incremental learning.

  Object detection problems are often formulated as classification tasks where a sliding window technique
  is used to scan the entire image and locate  objects of interest \cite{Papageorgiou2000Train,
  Viola2004Robust, Dalal2005HOG}. Viola and Jones \cite{Viola2004Robust} proposed an efficient
  detection algorithm based on the AdaBoost algorithm and a cascade classifier.
  Their detector is the first highly-accurate
  real-time face detector. They trained classifier on data sets with a few thousand 
  faces and a large
  number of negative non-faces. During the training procedure, negative samples are gradually
  bootstrapped and added to the training set of the boosting classifiers in the next stage. This
  method yields an extremely low false positive rate. A large number of faces and non-faces are used to
  cover different face appearances and poses, and the huge non-face possibilities.
  As a result, the computation cost and memory
  requirements for training an AdaBoost detector are unacceptably high. 
  Viola and Jones spent weeks on
  training a detector with $6,060$ features (weak learners) on a face training set of $4,916$.

  To speed up the training time bottleneck, a few approaches have been proposed.
  Pham and Cham \cite{Pham2007Fast} reduced the training time of weak learners 
  by approximating the decision
  stumps with class-conditional Gaussian distributions. Wu \etal 
  \cite{Wu2008Fast} introduced a fast
  implementation of the AdaBoost method and proposed forward feature selection (FFS) for fast training.
  FFS ignores the reweighting step in boosting such that 
  weak classifiers only need to be trained for once.
  Xiao \etal \cite{Xiao2007Dynamic} applied distributed learning to learn their proposed dynamic cascade
  framework. They use over $30$ desktop computers for parallel training. They managed to
  train a face detector on a training set with $500,000$ positive samples and $10$ billion
  negative samples in under $7$ hours. However, these techniques are not applicable to some
  real-world applications where a complete set of training samples is often not given in advance.
  Re-training the model each time new data arrive would increase the time 
  complexity by the factor of $N$,
  where $N$ is the number of newly arrived samples.
  Hence, developing an efficient adaptive object
  detector has become an urgent issue for many applications of object detection in diverse and
  changing environments. To alleviate this problem,
  online incremental learning algorithms have been
  proposed for this purpose.

  Online learning was firstly introduced in the 
  computational learning community. Since, boosting is one
  of the classifiers that have been successfully applied to many machine learning tasks, there has
  been considerable interest in applying boosting techniques on problems that require online
  learning.
  The first online version of boosting algorithms was proposed in
  \cite{Oza2001Online}. The algorithm works by minimizing the classification error while updating
  the weak classifiers online. Grabner and Bischof 
  \cite{Grabner2006Online} later applied online boosting to object
  detection and visual tracking. Based on 
  Oza and Russel's online boosting \cite{Oza2001Online}, 
  they proposed an online feature selection method, where a
  group of selectors is initialized randomly, each with its own feature pool. By interchanging weak
  learners based on lowest classification error, the algorithm is able to capture the change
  in patterns induced by new samples. Huang \etal 
  \cite{Huang2007Incremental} proposed an incremental learning
  algorithm that adjusts a boosted classifier with domain-partitioning weak hypotheses to online
  samples. They showed that by incremental learning with few difficult unseen faces (\eg,
  faces with sun glasses or extreme illumination), the performance of the online detector can be 
  significantly improved. Parag \etal \cite{Parag2008Boosting} advocated an online boosting algorithm
  where the
  parameters of the weak classifiers are updated using weighted linear regressor to minimize the
  weighted least square error. in the context of pedestrian detection,
  Liu and Yu \cite{Liu2007Gradient} introduced
  a gradient-based feature selection
  approach where the parameters of the weak classifiers are 
  updated using gradient descent to
  minimize weighted least square error. 
  %
  %
  Nonetheless, most of these proposed techniques concentrated
  on the application of visual tracking or object classification with small training sets and few
  online data sets\footnote{For example, 
  in \cite{Liu2007Gradient}, the authors
  trained the initial classifier with $366$ positive samples and 
  $2,540$ negative patches, and incrementally updated with $366$ online
  positive samples and $2,450$ online negative patches.}. Hence, to date,
  it remains unclear whether there is any improvement in object detection
  by continuously updating the existing models with a sufficiently large training sample set. 
  We will reveal this mystery 
  in Section~\ref{sec:cascade_gslda}.

  Recently, Moghaddam \etal \cite{Moghaddam2007Fast} presented a technique that combines the
  greedy approach with the
  efficient block matrix inverse formula. The proposed technique, termed greedy sparse linear
  discriminant analysis (GSLDA), speeds up the calculation time by $1000\times$ compared with
  globally optimal solutions found by branch-and-bound search
  in the case of binary-classification problems.
  Paisitkriangkrai \etal 
  \cite{Paisitkriangkrai2009CVPR} later
  applied the GSLDA algorithm to face detection and showed very convincing results. Their GSLDA
  face detector has shown to outperform AdaBoost based face detector due to the nature of the
  training data (the distribution of face and non-face samples is highly imbalanced). The objective
  of this work is to design an efficient incremental greedy sparse LDA algorithm that can
  accommodate new data efficiently while preserving a promising classification performance.

  Unlike classical LDA where a lot of online learning techniques have been designed and proposed
  \cite{Pang2005Incremental, Ye2005IDRQR, Zhao2008Incremental}, there are very few works on
  incremental learning for {\em sparse} LDA. One of the difficulties might be
  due to the fact that the
  sparse LDA problem is non-convex and NP-hard. 
  It is not straightforward to design an incremental solution for sparse LDA.
  In this work, we design an algorithm that efficiently learns and updates the
  sparse LDA classifier. Our online sparse LDA classifier not only incorporates new data efficiently
  but also yields an improvement in classification accuracy as new data become available. In brief,
  we extend the work of \cite{Paisitkriangkrai2009CVPR} with an efficient online update scheme.
  Our method modifies the weights of linear discriminant functions to adapt to new data sets. This
  update process generalizes the weights of linear discriminant functions and results in accuracy
  improvements on test sets.

  The key contributions of this work are summarized as follows.
  \begin{itemize}
    \item
    We propose an efficient incremental greedy sparse LDA classifier for training an object detector in
    an incremental fashion. The online algorithm integrates the GSLDA based feature selection with
    our adaptation schemes for updating weights of linear discriminant functions and the linear
    classifier threshold. Our updating algorithm is very efficient. We neither replace weak
    learners nor throw away any weak learners during updating phase.

    \item

%
    Our online GSLDA serves as a {\em better} (in terms of performance)
    alternative to the standard online boosting \cite{Oza2001Online} 
    for training
    detectors.     
    To our knowledge, it is the first time to apply the online sparse linear discriminant analysis
    algorithm to object detection. 

    \item
    Finally, we have conducted extensive experiments on several  data sets that have been used in
    the literature. The
    experimental results confirm that incremental learning with online samples is beneficial to the
    initial classifier. Our algorithm can efficiently update the classifier when the new instance
    is inserted while achieving comparable classification accuracy to the batch
    algorithm\footnote{We use the terms ``batch learning'' and ``offline learning'' interchangeably
    in this
    paper.}. Our
    findings indicate that online learning plays a crucial role in object detection, especially when
    the initial number of training samples is small. Note that when trained with few positive
    samples, the detector often {\em under-performs} since it fails to capture the appearance variations of
    the target objects. By applying our online technique, the classification performance can be
    further improved at the cost of a minor increase in training time.

  \end{itemize}

  The rest of the paper is organized as follows.  Section~\ref{sec:algo} begins by introducing the
  concept of LDA and GSLDA. We then propose our online GSLDA object detector. The results of
  numerous experiments are presented in Section~\ref{sec:exp}. We conclude the paper in
  Section~\ref{sec:conclusion}.

\begin{table}[t!]
  \caption{Notation}
  \begin{center}
    \begin{tabular}{c|l}
    \hline
    Notation & Description \\
    \hline
    \hline
      $C_1,C_2$  & Class $1$ (positive class), class $2$ (negative class) \\
      $N$        & Number of training samples in each classifier \\
                 & (cascade layer) \\
      $N_1,N_2$  & The number of training samples in first and second class,\\ 
                 & respectively \\
      $M$        & The size of the feature sets (for decision stumps, this is\\ 
                 & also equal to the number of weak learners) \\
      $T$        & The number of features to be selected \\
      $\bX$      & Data matrix \\
      $\bx$      & The new instance being inserted \\
      $\bar{\bm}$& The global mean of the training samples \\
      $\bm_1,\bm_2$& The mean (centroid) of the first and second class,\\
                 &   respectively \\
      $\Sigma_1,\Sigma_2$& The covariance of the first and second class \\
      $\mu_1,\mu_2$& The projected mean of the first and second class \\
      $\sigma_1,\sigma_2$& The projected covariance of the first and second class \\
      $S_b, \tilde{S}_b$ & Between-class scatter matrix and its updated value\\
                 &         after the new instance $\bx$ has been inserted\\
      $S_w, \tilde{S}_w$ & Within-class scatter matrix and its updated value \\
      $\bw$      & Weights of linear discriminant functions (also referred \\
                 & to as weak learners' coefficients in the context) \\
      $w_0$      & The linear classifier threshold \\
     \hline
    \end{tabular}
  \end{center}
  \label{tab:notations}
\end{table}

\section{Algorithms}
\label{sec:algo}

  For ease of exposition, the symbols and their denotations used in this paper are summarized in
  Table~\ref{tab:notations}. In this section, we begin by introducing the basic concept of classical
  linear discriminant analysis (LDA) and greedy sparse linear discriminant analysis (GSLDA).  We
  then propose our online greedy sparse LDA (OGSLDA).

\subsection{Classical Linear Discriminant Analysis}
\label{sec:lda}

  Linear discriminant analysis (LDA) deals with the problem of finding weights 
  of linear discriminant functions. Let us assume that
  we have a set of training 
  patterns $\bx = [x_{1},x_{2},...,x_{M}]^\T$ where each of which is assigned 
  to one of two classes, $C_{1}$ and $C_{2}$. We can find a weight vector 
  $\bw = [w_{1},w_{2},...,w_{M}]^\T$ and a threshold $w_{0}$ such that
  \begin{align}
  \label{EQ:Eq1}
    \bw^\T \bx + w_{0}  > 0  \quad (\bx \in C_{1}),  \nonumber\\
    \bw^\T \bx + w_{0}  < 0  \quad (\bx \in C_{2}).
  \end{align}

  In general, we seek the vector $\left[w_{0}, w_{1}, w_{2},..., w_{M}\right]$ 
  that best 
  satisfies \eqref{EQ:Eq1}. The data are said to be linearly separable if 
  for all $\bx$, \eqref{EQ:Eq1} is satisfied.
  
    An intuitive objective that one can take is to find a linear combination of the 
    variables that can separate the two classes as much as possible. The 
    computed linear combination reduces the dimensions of the samples to one 
    dimension. The classical criterion proposed by Fisher is the ratio of between-class 
    to within-class variances,
    which can be written as
    \begin{align}
      \label{EQ:J}
      J     &= \frac{\sum_{c \in C} \sum_{\bx \in c} 
            (\bw^\T\bm_{c}-\bw^\T\bar{\bm})^{2} }
            { \sum_{c \in C} \sum_{\bx \in c} 
            (\bw^\T \bx-\bw^\T \bm_{c})^{2} }
            = \frac{\bw^\T S_{b}\bw}{\bw^\T S_{w}\bw}, \\
      \label{EQ:Sb1}
      S_{b} &= \sum_{c \in C} 
            {N_{c}(\bm_{c}-\bar{\bm})(\bm_{c}-\bar{\bm})^\T}, \\
      \label{EQ:Sw1}
      S_{w} &=  \sum_{c \in C} \sum_{\bx \in c} 
            {(\bx-\bm_{c})(\bx-\bm_{c})^\T}.
    \end{align}

    Here, $\bm_{c}$ is the mean of class $c$, $\bar{\bm}$ is the global mean,
    $N_c$ is the number of instances in class $c$, $S_b$ and $S_w$ are
    the so-called between-class and within-class scatter matrices. The 
    numerator of \eqref{EQ:J} denotes the distance between the 
    projected means and the denominator denotes the 
    variance of the pooled data. We want to find 
    linear projections $\bw$ that maximizes $J$, the distance between the 
    means of the two classes while minimizing the variance within each class. 
    The solution can be obtained by generalized eigen-decomposition (GEVD).
    The optimal solution $\bw$ is the eigenvector corresponding to the maximal 
    eigenvalue and can be expressed as \cite{Duda00pattern}:
    \begin{align}
      \label{EQ:FDA}
      \bw \propto S^{-1}_{w}(\bm_{1}-\bm_{2}).
    \end{align}
    
    If we further assume that the data are normally distributed and that the 
    distributions in the original space have identical covariance matrices, an 
    optimal threshold, $w_0$, can be calculated from 
    \begin{align}
      \label{EQ:optThresh}
      w_{0} = -\frac{1}{2} (\bm_{1}+\bm_{2})^\T S^{-1}_{w} (\bm_{1}-\bm_{2})
              -\log \left(\frac{\Pr(C_{2})}{\Pr(C_{1})} \right).
    \end{align}
    
    Here, $\Pr(C_1)$ and $\Pr(C_2)$ are priori probabilities of class $C_1$ and 
    $C_2$, respectively. This threshold can be interpreted as the mid-point 
    between the two projected means, shifted by the log of the ratio between 
    the priori probabilities of the two classes.
 
  %
  %
  \subsection{Greedy Sparse Linear Discriminant Analysis}
  
  In this section, we briefly present the offline implementation of the 
  greedy sparse LDA 
  algorithm \cite{Moghaddam2006Generalized, Moghaddam2007Fast}. The sparse 
  version of classical LDA is to solve 
  \begin{align}
    \maximize_{\bw} \quad & \frac{ \bw^\T S_b \bw }{ \bw^\T S_w \bw } , \\
    \st        \quad & \card(\bw) = k, \notag
  \end{align}
  where $\card(\bw) = k$ is an additional sparsity constraint, $\card(\cdot)$ 
  denotes $\ell_0$ norm, $k$ is an integer set by a user. 
  Due to this additional 
  sparsity constraint, the problem is non-convex and NP-hard. 
  In \cite{Moghaddam2006Generalized}, Moghaddam \etal presented a technique 
  to compute optimal sparse linear discriminants using branch-and-bound 
  approach. Nevertheless, finding the exact global optimal solutions 
  for high dimensional data is infeasible. The algorithm was extended in 
  \cite{Moghaddam2007Fast} with new sparsity bounds and efficient block matrix 
  inverse techniques to speed up the computation time by $1000\times$. 
  The technique works by sequentially adding the new variable which yields the 
  maximum eigenvalue (forward selection) until the number of 
  nonzero components, $\card(\bw)$, is equal to the integer set by the user.

   In \cite{Paisitkriangkrai2009CVPR}, Paisitkriangkrai \etal
   learn an object 
   detector using GSLDA algorithm. The training procedure is described in 
   Algorithm~\ref{ALG:GSLDA}. 
   First, the set of selected features is initialized to an empty set.
   The algorithm then trains all weak learners and store their results 
   into a lookup table (line $1 - 2$). At every 
   round, the output of each weak learner is examined and the weak learner 
   that most separates the two classes is sequentially added to the list
   (line $3 - 4$). 
   Mathematically, Algorithm~\ref{ALG:GSLDA} sequentially selects the weak 
   learner whose output yields the maximal 
   eigenvalue. Weak learners are added until the target learning goal is met. 
   The authors of \cite{Paisitkriangkrai2009CVPR}
   use an asymmetric node 
   learning goal to build a cascade of GSLDA object detector.

%
%
%
%
\SetVline
\linesnumbered

\begin{algorithm}[t]
\caption{The training procedure for building an offline GSLDA object detector.}
\begin{algorithmic}
\small{
   \KwIn{
   \begin{itemize}
      \item
         A positive training set and a negative training set;
      \item
         A set of features $\left\{f_i; i \in [1, M]\right\}$;
   \end{itemize}
   }

   \KwOut{
\begin{itemize}
   \item
      A set of selected weak learners $\left\{h_i; i \in [1, T]\right\}$ 
      that best separates the training set;
\end{itemize}
}

\ForEach{feature}
{
  Train a weak learner (\eg,
  a decision stump parameterized by a threshold        
  %
  %
  that results in the smallest classification error) 
  on the training set;
}

\While{ the target goal is not met }
{
  Add the best weak learner (\eg, decision stump) that yields 
  the maximum class separation to the set of selected weak learners;
}

} 
\end{algorithmic}
\label{ALG:GSLDA}
\end{algorithm}

  \subsection{Incremental Learning of GSLDA Classifiers}
        
  The major challenge of GSLDA object detectors in real-world applications is that a complete set of
  training samples is often not given in advance. As new data arrive, the between-class and
  within-class scatter matrices, $S_b$ and $S_w$, will change accordingly. In offline GSLDA, the
  value of both matrices would have to be recomputed from scratch. However, this approach is
  unacceptable due to its heavy computation and storage requirements. First, the cost of computing
  both matrices grows with the number of training samples. As a result, the algorithm will run
  slower and slower as time progresses.
  Second, the batch approach uses the entire set of training data for
  each update. In other words, the previous training data needs to be stored for the
  retraining purpose.

  In order to overcome these drawbacks, we propose an online learning algorithm, 
  termed online greedy
  sparse LDA (OGSLDA). The OGSLDA algorithm consists of two phases: the initial offline learning
  phase and the incremental learning phase. The training procedure in the initial phase is similar
  to the algorithm outlined in Algorithm~\ref{ALG:GSLDA}. Here, we assume that the number of
  training samples available initially is adequate and well represents the true density. In the
  second phase, the learned covariance matrices are updated in an incremental manner.

  It is important to point out that a number of incremental LDA-like approximated algorithms have been
  proposed in \cite{Ye2005IDRQR,Kim2007Incremental}. 
  Ye \etal \cite{Ye2005IDRQR} proposed
  an efficient LDA-based incremental dimension reduction algorithm which applied QR decomposition
  and QR-updating techniques for memory and computation efficiency. 
  Kim \etal \cite{Kim2007Incremental}
  proposed an incremental LDA by applying the concept of the sufficient spanning set approximation
  in each update step. However, we did not find any of the existing LDA-like algorithms appropriate
  to our problems. Based on our preliminary experiments, the projection matrix determined in
  subspace often gives worse discriminant power than that from full space.
  %
  %
  This might be due to their dimension reduction algorithms which reduced
  between-class and within-class scatter matrices to a much smaller size.
  Our online GSLDA guaranteed to build the same between-class and within-class
  scatter matrices as batch GSLDA given the same training data.
  The reason why we need not worry about large dimensions in our algorithms 
  is because
  applying sparse LDA in our initial phase already reduces the number 
  of dimensions we have to deal with. 
  Hence, given the same set of features, the accuracy of our online GSLDA is 
  better than the existing incremental LDA-like approximated algorithms.
  The only expensive computation left in our algorithms is eigen-analysis. In order to avoid the high computation complexity of
  continuously solving generalized eigen-decomposition, we applied the efficient matrix inversion
  updating techniques based on inverse Sherman-Morrison formula. As a result, our incremental
  algorithm is very robust and efficient.

  In this section, we first introduce an efficient method that incrementally updates both
  within-class and between-class scatter matrices as new observations arrive. Then, an approach used
  to update the classifier threshold is described. Finally, we analyze the storage and training time
  complexity of the proposed method.

  \subsubsection{Incremental update of between-class and within-class matrices}

    Since, GSLDA assumes Gaussian distribution, the incremental update of class mean and class
    covariance can be computed very quickly. The techniques used to update both matrices can be
    easily derived. The procedure proceeds in three steps:
    
    \begin{enumerate}
      \item Updating between-class scatter matrix, $S_b$;
      \item Updating within-class scatter matrix, $S_w$;
      \item Updating inverse of within-class scatter matrix, $S_w^{-1}$.
    \end{enumerate}
    
    \emph{Updating between-class scatter matrix:}
    The definition of the between-class scatter matrix is given in \eqref{EQ:Sb1}. For $2$ classes
    ($C_1$ and $C_2$), $S_b$ can be simplified to 
    \begin{align}
      \label{EQ:Sb2}
      S_b = \frac{N_1 N_2}{N} (\bm_1 - \bm_2)(\bm_1 - \bm_2)^\T.
    \end{align}
    The expression can be interpreted as the scatter of class $1$ with respect to the scatter of
    class $2$. Let $\bx$ be a new instance being inserted. The updated $\tilde{\bm}_1$ and
    $\tilde{\bm}_2$ can be calculated from
    \begin{align}
       \label{EQ:mtilde}
       \tilde{\bm_{1}} =
        \begin{cases}
          \bm_{1} + \frac{\bx - \bm_{1}}{N_1 + 1} & \textrm{if } \bx \in C_1;\\
          \bm_{1} & \textrm{otherwise, }
        \end{cases} \nonumber\\
       \tilde{\bm_{2}} =
        \begin{cases}
          \bm_{2} & \textrm{if } \bx \in C_1;\\
          \bm_{2} + \frac{\bx - \bm_{2}}{N_2 + 1} & \textrm{otherwise. }
        \end{cases}
    \end{align}

    \emph{Updating within-class scatter matrix:}
    The covariance of a random vector $\bX$ is a square matrix $\Sigma$ where $\Sigma = \bE[(\bX -
    \bE[\bX])(\bX - \bE[\bX])^\T]$. Given the new instance $\bx$, the updated covariance matrix is
    given by
    \begin{equation}
      \label{EQ:sigmaTilde}
      \tilde{\Sigma} = ([\bX, \bx] - \tilde{\bm} \b1^\T)
      ([\bX, \bx] - \tilde{\bm} \b1^\T)^\T.
    \end{equation}
    Here, $\tilde{\bm}$ is an updated mean after new instance has been inserted and $\b1$ is a
    column vector with each entry being $1$. Its dimensionality should be clear from
    the context. 
    Note that in \eqref{EQ:sigmaTilde}, we leave out the
    constant term since it makes no difference to the final solution:
    \begin{align}
     [\bX, \bx] - \tilde{\bm} \b1^\T 
     &= [\bX, \bx] - \bm \b1^\T + \bm \b1^\T - \tilde{\bm} \b1^\T \nonumber \\
     %
     &= [\bX-\bm \b1^\T, \bx - \bm] - (\tilde{\bm} - \bm) \b1^\T. \nonumber 
    \end{align}
    Substitute the above expression into \eqref{EQ:sigmaTilde} and
    let $\bu = \bx - \bm$ and $\bv = \tilde{\bm} - \bm$,
    \begin{align}
      \label{EQ:sigmaTilde2}
      \tilde{\Sigma} &= ([\bX-\bm \b1^\T, \bu] - \bv \b1^\T)
       ([\bX-\bm \b1^\T, \bu] - \bv \b1^\T)^\T \nonumber\\
       &= ([\bX-\bm \b1^\T, \bu][\bX-\bm \b1^\T, \bu]^\T - 
       [\bX-\bm \b1^\T, \bu] (\bv \b1^\T)^\T \nonumber\\
       &\qquad- (\bv \b1^\T)[\bX-\bm \b1^\T, \bu]^\T +
       (\bv \b1^\T)(\bv \b1^\T)^\T  \nonumber\\
       &= \Sigma + \bu \bu^\T -
       [\bX-\bm \b1^\T, \bu] \b1 \bv^\T  \nonumber\\
       &\qquad- \bv ([\bX-\bm \b1^\T, \bu] \b1)^\T +
       (N+1)\bv \bv^\T \nonumber\\
       &= \Sigma + \bu \bu^\T -
       (N\bm-N\bm+\bu) \bv^\T  \nonumber\\
       &\qquad- \bv (N\bm-N\bm+\bu)^\T +
       (N+1)\bv \bv^\T \nonumber\\
       &= \Sigma + \bu \bu^\T - \bu \bv^\T -\bv\bu^\T +
       (N+1)\bv \bv^\T \nonumber\\
       &= \Sigma + (\bu - \bv)(\bu - \bv)^\T + N \bv \bv^\T \nonumber\\
       &= \Sigma + (\bx - \tilde{\bm}) (\bx - \tilde{\bm})^\T 
       + N(\tilde{\bm} - \bm)(\tilde{\bm} - \bm)^\T.
    \end{align}

    Note that $\bX \b1 = \bm \b1^\T \b1 = N \bm$.
    Next, we consider updating within-class scatter matrix. Let $\bx$ be a new instance being
    inserted. The updated matrix, $\tilde{S}_w$, can be calculated from
     \begin{align}
      \tilde{S}_w =
        \begin{cases}
          \tilde{\Sigma}_{1} + \Sigma_{2} & \textrm{if } \bx \in C_1;\\
          \Sigma_{1} + \tilde{\Sigma}_{2} & \textrm{otherwise. }
        \end{cases}
    \end{align}
    
    \emph{Updating inverse of within-class scatter matrix:}
    As mentioned in \cite{Moghaddam2007Fast} that the computational complexity of $2$-class GSLDA
    relies heavily on the calculating of within-class scatter matrix inversion. In order to update
    the matrix inversion efficiently, we make use of the technique called Balanced Incomplete
    Factorization which was based on inverse Sherman-Morrison formula proposed by Bru \etal in
    \cite{Bru2008Balanced}. Let $\Sigma$ be the square matrix of size $M \times M$ which can be
    written as 
    \begin{align}
      \label{EQ:sigma2}
      \Sigma = \Sigma_0 + \bp_1 \bq_1^\T + \bp_2 \bq_2^\T.
    \end{align}
    
    Here, we assume that $\Sigma_0$ is nonsingular and $\bp_1,\bp_2,\bq_1,\bq_2 \in \mathbb{R}^M$.
    The inverse of $\Sigma$ is given by 
    \begin{align}
      \label{EQ:BIF}
      \Sigma^{-1} = \Sigma_0^{-1} - \Sigma_0^{-1} U D^{-1} V^\T \Sigma_0^{-1}
    \end{align}
    where 
    $D^{-1} = \begin{bmatrix}
     r_1^{-1}  & 0        \\
     0         & r_2^{-1}
    \end{bmatrix}$,
    $U = \begin{bmatrix}
    \bp_1  & \bp_2 - \frac{\bq_1 \Sigma_0^{-1} \bp_2}{r_1} \bp_1
    \end{bmatrix}$, 
    $V = \begin{bmatrix}
    \bq_1  & \bq_2 - \frac{\bq_2^\T \Sigma_0^{-1} \bp_1}{r_1} \bq_1
    \end{bmatrix}$, 
    $r_1 = 1 + \bq_1 \Sigma_0^{-1} \bp_1$, 
    \[
            r_2 = 1 + \left(
                     \bq_2 - \frac{\bq_2^\T \Sigma_0^{-1} \bp_1}{r_1} \bq_1
                     \right) 
                     ^\T 
           \Sigma_0^{-1} \bp_2.
    \]
           
    The updated inverse of within-class scatter matrix can be written as
    \begin{align}
      \label{EQ:Sw_inv}
      \tilde{S}_w^{-1} = S_w^{-1} - S_w^{-1} U D^{-1} V S_w^{-1}
    \end{align}
    where $\bp_1 = \bq_1 = \bx - \tilde{\bm}$, $\bp_2 = N_c(\tilde{\bm} - \bm)$ and $\bq_2 =
    \tilde{\bm} - \bm$ (from \eqref{EQ:sigmaTilde2} and \eqref{EQ:sigma2}).

    \subsubsection{Updating weak learners' coefficients and threshold} 
    \label{sec:OGSLDA_thresh}
    
    Given the updated within-class matrix, $\tilde{S}_w^{-1}$, and between-class matrix,
    $\tilde{S}_b$, the updated weights of linear discriminant functions can now be calculated from
    matrix-vector multiplication using \eqref{EQ:FDA}. To complete the linear classifier, the
    threshold $w_0$ has to be obtained. Three criteria can be adopted. The first criterion is to
    apply the optimal Bayesian classifier in the projected space. In other words, the selected
    threshold should be the value in which the one-dimensional distribution functions in the
    projected lines are equal. The mean and variance in the transformed space can be calculated as
    \begin{align}
      \mu_c = \bw^\T \bm_c, 
      \quad
      \sigma_c = \bw^\T \Sigma_c \bw. 
    \end{align}
    
     If we let $X_1 \sim \cN(\mu_1,\sigma_1^2)$ and $X_2 \sim \cN(\mu_2,\sigma_2^2)$. The optimal
     threshold is calculated as the point in which the one-dimensional density function of two
     classes are equal. Let $\log \Pr(x_1) = \log \Pr(x_2)$. After some algebraic expansions and
     simplifications, we can write the expression in the second-order polynomial,
     \begin{align}
       a x^2 + b x + c = 0 \nonumber
     \end{align}
     where $a = -\frac{1}{2 \sigma_1^2} + \frac{1}{2 \sigma_2^2}$,
     $b = \frac{\mu_1}{\sigma_1^2} - \frac{\mu_2}{\sigma_2^2}$ and
     $c = \frac{\mu_2^2}{2 \sigma_2^2} + \log(\sigma_2) -
     \frac{\mu_1^2}{2 \sigma_1^2} + \log(\sigma_1)$. 
     The quadratics have two roots,
     \begin{align}
       x = \frac{-b \pm \sqrt{b^2 - 4 a c}}{2a}. \nonumber
     \end{align}
     
     In our implementations, we choose the threshold, $w_0$, to be the value between the
     two class means,
     \begin{align}
       \label{EQ:w0Crit1}
       w_0 = x \textrm{ where } \mu_1 < x < \mu_2.
     \end{align}
     
    The second criterion is to choose the threshold which yields high detection rate with moderate
    false alarm rate. This asymmetric criterion is often adopted in cascade framework
    \cite{Viola2004Robust}. Let $\phi(Z) = \frac{1}{\sqrt{2 \pi}}
    \int_{-\infty}^{Z}{\exp(-\frac{1}{2}u^2)} du$ be the cumulative distribution function (CDF) of
    the standard normal random variable $Z$. If $X \sim \cN(\mu_1,\sigma_1^2)$, the CDF of $X$ is
    $\phi(Z)$ where $Z = \frac{X - \mu_1}{\sigma_1}$. Let the miss rate by $p$, the threshold which
    yields $1 - p$ detection rate can be calculated as 
    \begin{align}
      \label{EQ:w0Crit2}
      w_0 = \mu_1 + Z \sigma_1 = \mu_1 + \phi^{-1}(p) \sigma_1.
    \end{align}
    
    The last criterion is to set the threshold to be the projected mean of the negative classes.
    This threshold helps us ensure the target asymmetric learning goal (moderate ($50\%$) false
    positive rate with high detection rate). The threshold for the last criterion is
    \begin{align}
      \label{EQ:w0Crit3}
      w_0 = \mu_2.
    \end{align}
    
    The above three threshold updating rules might look oversimple. However,
    in \cite{Rueda2004Efficient}, a few  numerical simulations were performed
    on multi-dimensional normally
    distributed classes and real-life data taken form UCI machine learning repository. It is
    reported that selecting threshold using the simple approach as 
    \eqref{EQ:w0Crit1} often leads to smaller
    classification error than the traditional Fisher's approach \eqref{EQ:optThresh}.

\begin{figure*}[tb!]
  \centering
    \includegraphics[width=0.85\textwidth]{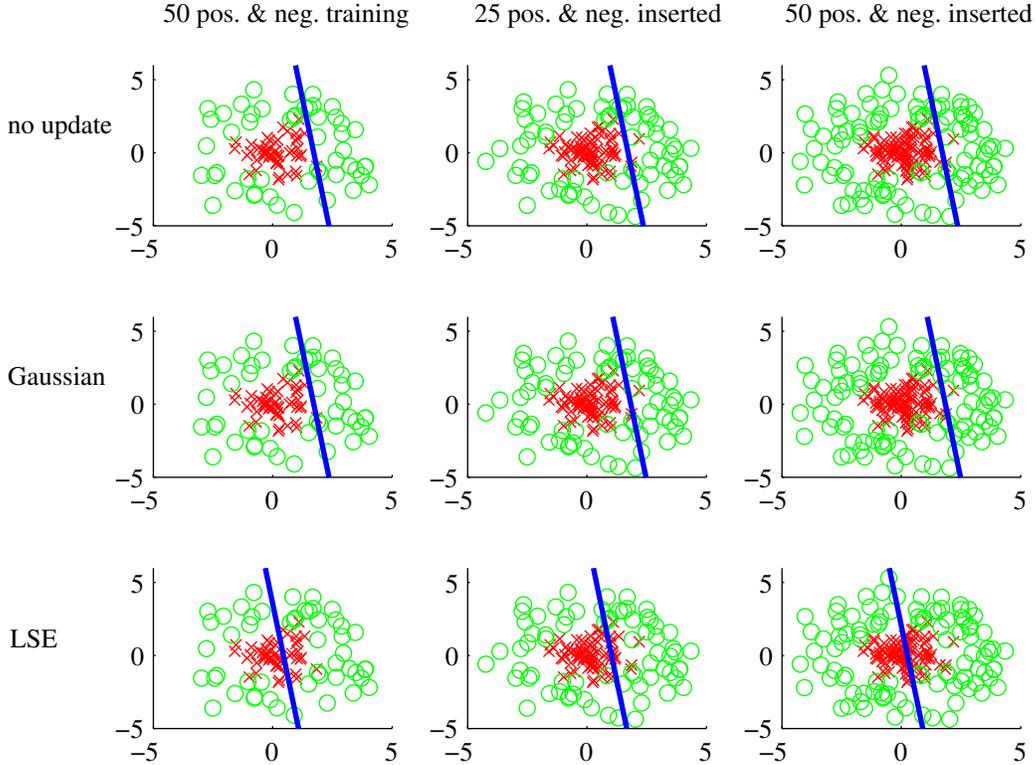}
\caption{Toy data set. $\times$'s and $\circ$'s represent positive and negative samples, respectively. {\bf
Top row:} No update. The parameters of weak learners do not get updated. {\bf Middle row:} Gaussian model.
Linear classifier threshold is calculated from updated mean and variance (using \eqref{EQ:w0Crit1}).
{\bf Bottom row:} Least square error. Linear classifier threshold is updated using linear regression. 
The leftmost column shows the classifier thresholds on the initial training set ($ 50 $ positive and $
50 $ negative training points). The middle column shows the classifier thresholds with $ 25 $ 
new positive and $ 25 $ new negative points inserted. The rightmost column shows the thresholds with 
$ 50 $ new positive and $ 50 $ new negative inserted. 
Due to the asymmetry of the data distributions, updating the parameters of the weak learners could
result in performance deterioration.
}
 \label{fig:compare_linear}
\end{figure*}

    Unlike many online boosting algorithms which modify the parameters of the weak learners to adapt
    to new dataset. For example, in \cite{Grabner2006Online},
    the parameters of the weak learners are updated using
    Kalman filtering;
    Parag \etal \cite{Parag2008Boosting} updated the parameters using linear regression;
    Liu and Yu \cite{Liu2007Gradient} updated the parameter using gradient descent, \etc.
    We have found that extreme
    care has to be taken when we consider updating weak learners' parameters for application of
    object detection.
    To demonstrate this, we generate an artificial asymmetric data set similar to
    the one used in \cite{Viola2002Fast}. 
    We then learn two different incremental linear weak classifiers
    with different parameter updating schemes: 
    \begin{enumerate}
        \item
    Incrementally update the model based on
    Gaussian distribution similar to \cite{Grabner2006Online};
        \item
    Incrementally update linear
    coefficients and intercept to minimize least square error (LSE) using linear regression similar
    to \cite{Parag2008Boosting} (here, we assume uniform sample weights).
    \end{enumerate}
    In this experiment, each
    weak learner represents a linear function with different coefficients (slopes). Each weak
    learner has one updatable parameter, \ie, linear classifier threshold (intercept). 
    We apply GSLDA
    algorithm and select the weak learner with minimal classification error. Based on the selected
    weak learner, we continuously insert new samples and update the linear classifier threshold.
    Fig.~\ref{fig:compare_linear} plots $ 9 $ different linear classifier thresholds. 
    Top row shows
    the linear classifier with no parameter updating. Middle row shows the linear classifier with
    Gaussian updating rule. Bottom row shows the linear classifier using 
    the linear regression algorithm.
    The first column shows the classifier thresholds on the initial training set. The middle and
    last columns show the classifier thresholds with new data being inserted. 
    We found that the top two
    classifier thresholds (no update and Gaussian) perform very similarly. LSE seems to perform worse
    when more new data are inserted. The reason may be attributed to the asymmetry of the data. When the
    data are linearly separable, we can see that the regressor works very well. 
    Based on our
    results, we feel that parameter updating algorithms could significantly weaken the performance of
    weak learners if not applied properly. Hence, in this work, we decide
    not to update the parameters of the
    weak learners in our algorithms.
    Clearly, another benefit is faster computation with no updating the weak learners'
    parameters.

%
%

    The online
    GSLDA framework is summarized in Algorithm~\ref{ALG:OGSLDA}.
    Note that here we only use forward search of the
    original GSLDA algorithm of \cite{Moghaddam2006Generalized,Moghaddam2007Fast}.
    In \cite{Paisitkriangkrai2009CVPR}, we have shown that forward selection plus backward
    elimination improve the detection performance {\em slightly} but with extra computation.

%
%
%
%
\SetKwInput{KwGiven}{Given}

\SetVline
\linesnumbered

\begin{algorithm}[t]
\caption{The online GSLDA Algorithm.}
\begin{algorithmic}
\small{
  \KwGiven{
   \begin{itemize}
      \item
         The initial set of weak learners $\left\{h_i; i \in [1, T]\right\}$ 
         trained using offline GSLDA on small initial data;
   \end{itemize}
   }

   \KwIn{
   \begin{itemize}
      \item
         New training datum $I$ and its corresponding class 
         label $y \in \left\{1,2\right\}$;
      \item
         The current between-class covariance matrix, $S_b$;
      \item
         The inverse of within-class covariance matrix, $S_w^{-1}$;
   \end{itemize}
   }

   \KwOut{
\begin{itemize}
      \item
         The updated between-class covariance matrix, $\tilde{S}_b$;
      \item
         The updated inverse of within-class covariance matrix, 
         $\tilde{S}_w^{-1}$;
      \item
         The updated weak learners' coefficients, $\tilde{\bw}$;
      \item
         The classifier threshold, $\tilde{w}_0$
   \end{itemize}
   }
   
   Classify the new datum $I$ using the given weak learners, 
      $\bx = [h_1(I), h_2(I), \cdots, h_T(I)]$; \\
   Update $S_b$ with $\bx$ using \eqref{EQ:Sb2} and~\eqref{EQ:mtilde}; \\
   Update $S_w^{-1}$ using \eqref{EQ:Sw_inv}; \\
   Recalculate weak learners' coefficients, $\bw$, using \eqref{EQ:FDA}; \\
   Update classifier threshold, $w_0$, based on node learning goal 
   (\eqref{EQ:w0Crit1} for minimal classification error, or
   $\min($\eqref{EQ:w0Crit2}$,$\eqref{EQ:w0Crit3}$)$ for the asymmetric
   node learning goal  (see Section~\ref{sec:OGSLDA_thresh}));
} 
\end{algorithmic}
\label{ALG:OGSLDA}
\end{algorithm}

    \subsubsection{Incremental Learning Computational Complexity}
    \label{sec:complexity}

    Since, the initial training of online GSLDA is the same as offline GSLDA, we briefly explain the
    time complexity of GSLDA \cite{Paisitkriangkrai2009CVPR}. Let us assume we choose decision
    stumps as our weak learners. Let the number of training samples be $ N $.
    Finding an optimal
    threshold of each feature needs $O(N \log{N}$).\footnote{One usually sorts the 1D features using
    Quicksort, which
    has a complexity $O(N \log{N})$.}
    Assume that the size of the feature set is $M$.
    The time complexity for training weak learners is $O(MN\log{N})$. During GSLDA learning, we need
    to find mean $O(N)$, variance $O(N)$ and correlation $O(T^2)$ for each feature. Since, we have
    $M$ features and the number of weak learners to be selected is $T$, the total time complexity
    for offline GSLDA is $O(MN\log{N} + MNT + MT^3)$.

    Given the selected set of weak learners, the time complexity of online GSLDA when new instance
    is inserted can be calculated as follows. Since, the number of weak learners is $T$, the total
    time complexity to calculate $\bx$ in Step $1$ is $O(T)$. It also takes $O(T)$ to update the class
    mean in Step $2$. At Step $3$, calculating $U$, $V$, $r1$, $r2$ take $O(T^2)$. In this step, the
    order in which we calculate the matrix-matrix multiplication affects the overall efficiency.
    Since, we are dealing with a small matrix chain multiplication, it is possible to go through
    each possible order and pick the most efficient one. For \eqref{EQ:BIF}, we perform
    matrix-matrix multiplication in the following order $ (((\Sigma_0^{-1} U) D^{-1}) (V^\T
    \Sigma_0^{-1})) $. The number of operations required to compute $(\Sigma_0^{-1} U)$ is $O(T
    \times T \times 2)$, $((\Sigma_0^{-1} U) D^{-1})$ is $O(T \times 2 \times 2)$, $(V^\T
    \Sigma_0^{-1})$ is $O(2 \times T \times T)$ and $(((\Sigma_0^{-1} U) D^{-1}) (V^\T
    \Sigma_0^{-1}))$ is $O(T \times 2 \times T)$. Hence, the complexity of updating matrix inversion
    is still in the order of $O(T^2)$. Since, the size of within-class matrix is $T \times T$, the
    matrix-vector multiplication in Step $4$ takes $O(T^2)$. Updating classifier threshold in Step
    $5$ takes $O(T^2)$ for the first criterion (First, we find the projected mean and covariance,
    $O(T)$ and $O(T^2 + T)$, respectively. Then, we solve a
    closed-form second-degree polynomial). The
    second criterion in Step $5$ takes $O(T^2)$  (Again, the time complexity of projected mean and
    covariance is $O(T)$ and $O(T^2 + T)$). The third criterion in Step $5$ takes $O(T)$ (Here, we
    only have to calculate the dot product of two vectors). Hence, the time complexity of Step $5$
    is at most $O(T^2)$. Therefore, the total time complexity for online GSLDA with the insertion of
    a new instance is at most $O(\underbrace{N_0M \log N_0 + N_0 MT + MT^3}_{\text{Offline}} +
    \underbrace{T^2}_{\text{Online}})$.  Here, $N_0$ is the number of initial training samples which
    assumed to be small. Note that the speed-up of online GSLDA over batch GSLDA is noticeable,
    \ie $\underbrace{O(NT^2)}_{\text{Online}}$
    $\ll$
    $\underbrace{O(N^2\log{N})}_{\text{Batch}}$,
    when more instances are inserted into the training set ($N \gg N_0$).

In terms of memory usage, between-class scatter matrix takes up $O(2T)$. The inverse of within-class
scatter matrix occupies $O(T^2)$. For the first and second criteria in Step $5$, we also need to
keep the covariance matrices of $\Sigma_1$ and $\Sigma_2$ which takes up $O(2T^2)$. Hence, the extra
memory requirements for online GSLDA are at most $O(3T^2 + 2T)$. Given that the selected number of weak
classifiers in each cascade layer is often small ($T < 200$), the time and memory complexity
of online GSLDA is almost negligible.

    

\section{Experiments}
\label{sec:exp}

This  section is organized as follows. The datasets used in this experiment, including how the
performance is analyzed, are described. Experiments and the parameters used are then discussed.
Finally, experimental results and analysis of different techniques are presented.

  \subsection{USPS Digits Classification}
  \label{sec:USPS}

  \begin{figure*}[tbh!]
    \begin{center}
      \subfigure[]
      {
        \includegraphics[width=0.30\textwidth,clip]{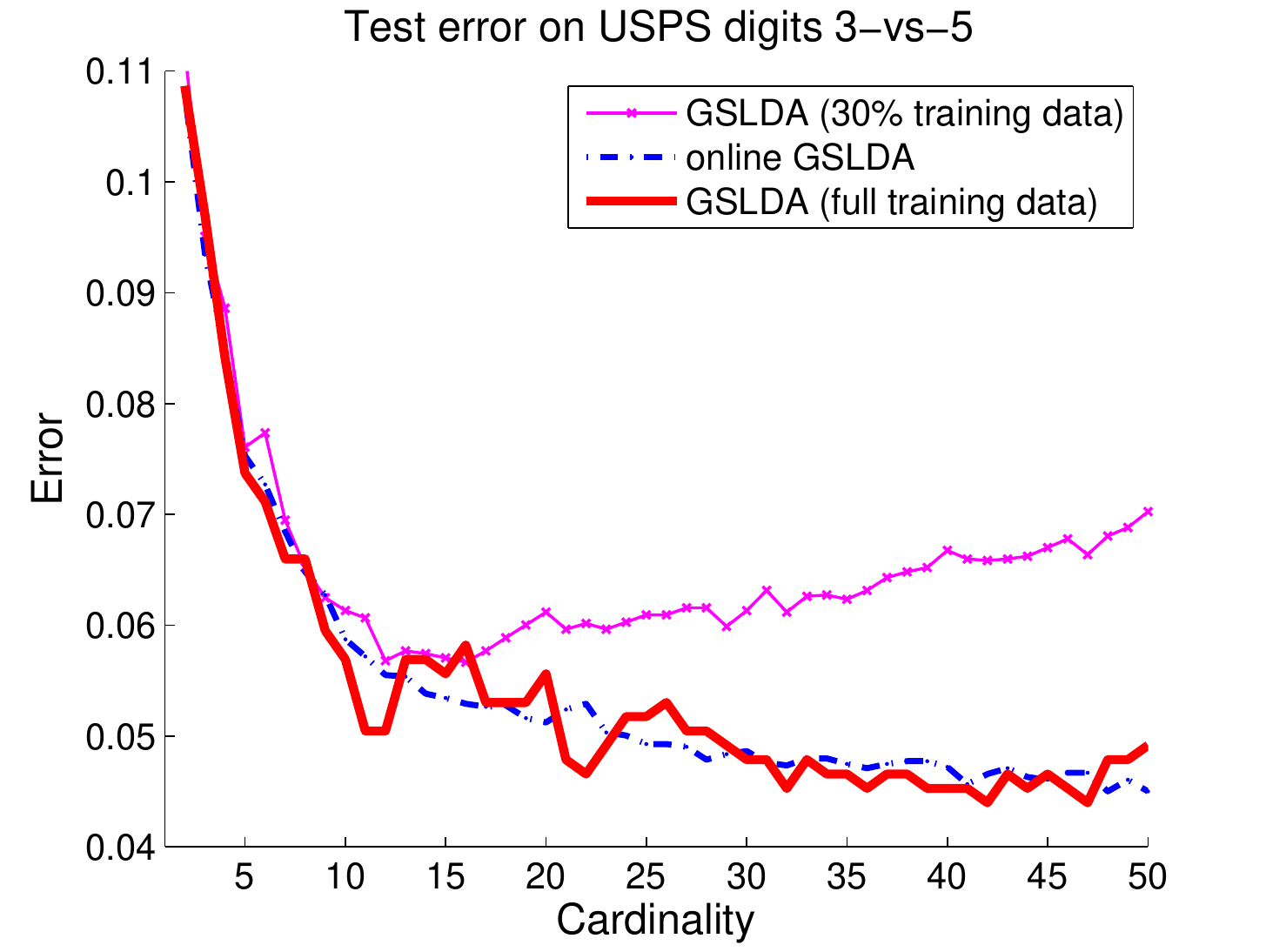}
        \label{fig:USPS1a}
      }
      \subfigure[]
      {
        \includegraphics[width=0.30\textwidth,clip]{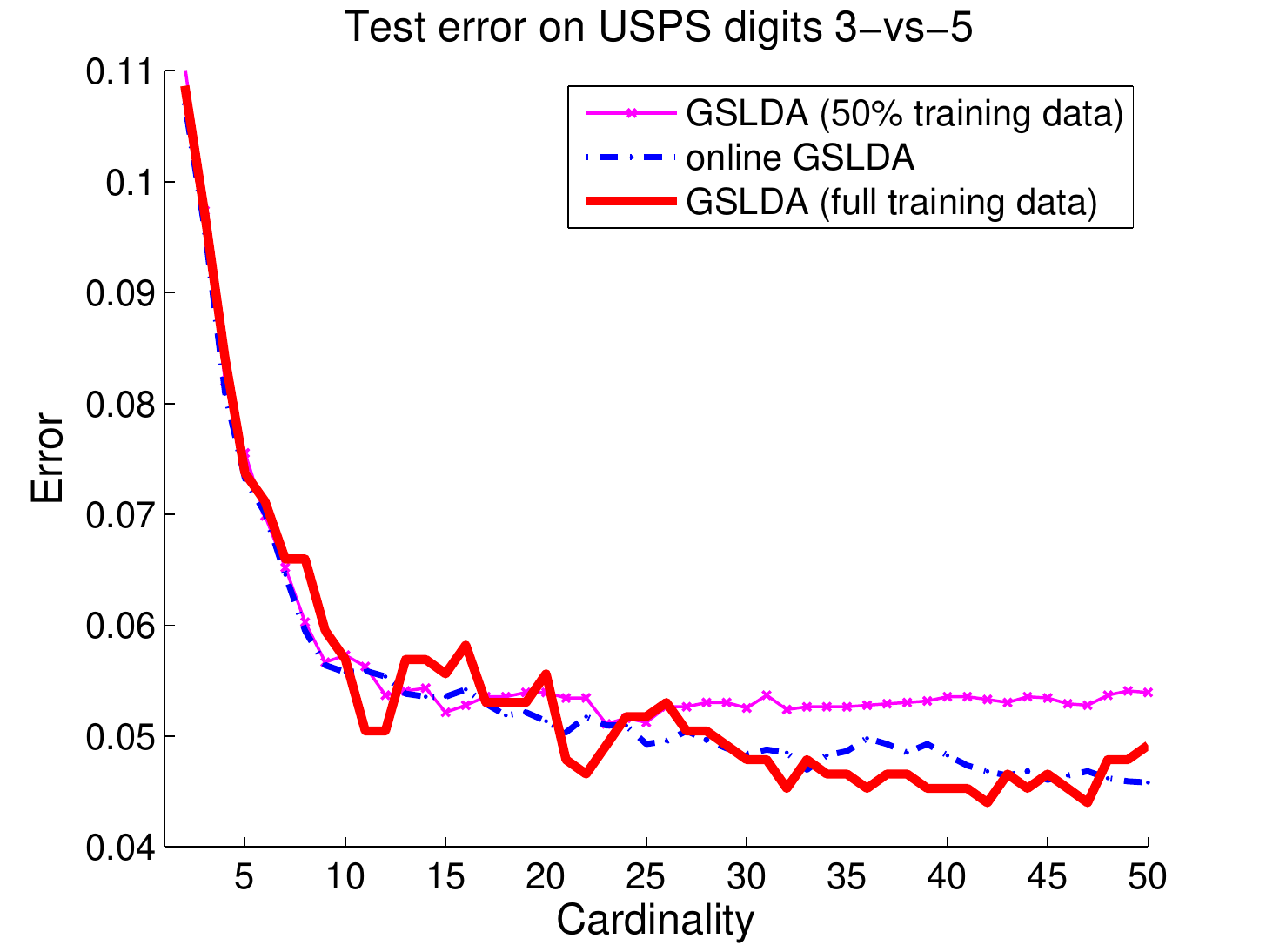}
        \label{fig:USPS1b}
      }
      \subfigure[]
      {
        \includegraphics[width=0.30\textwidth,clip]{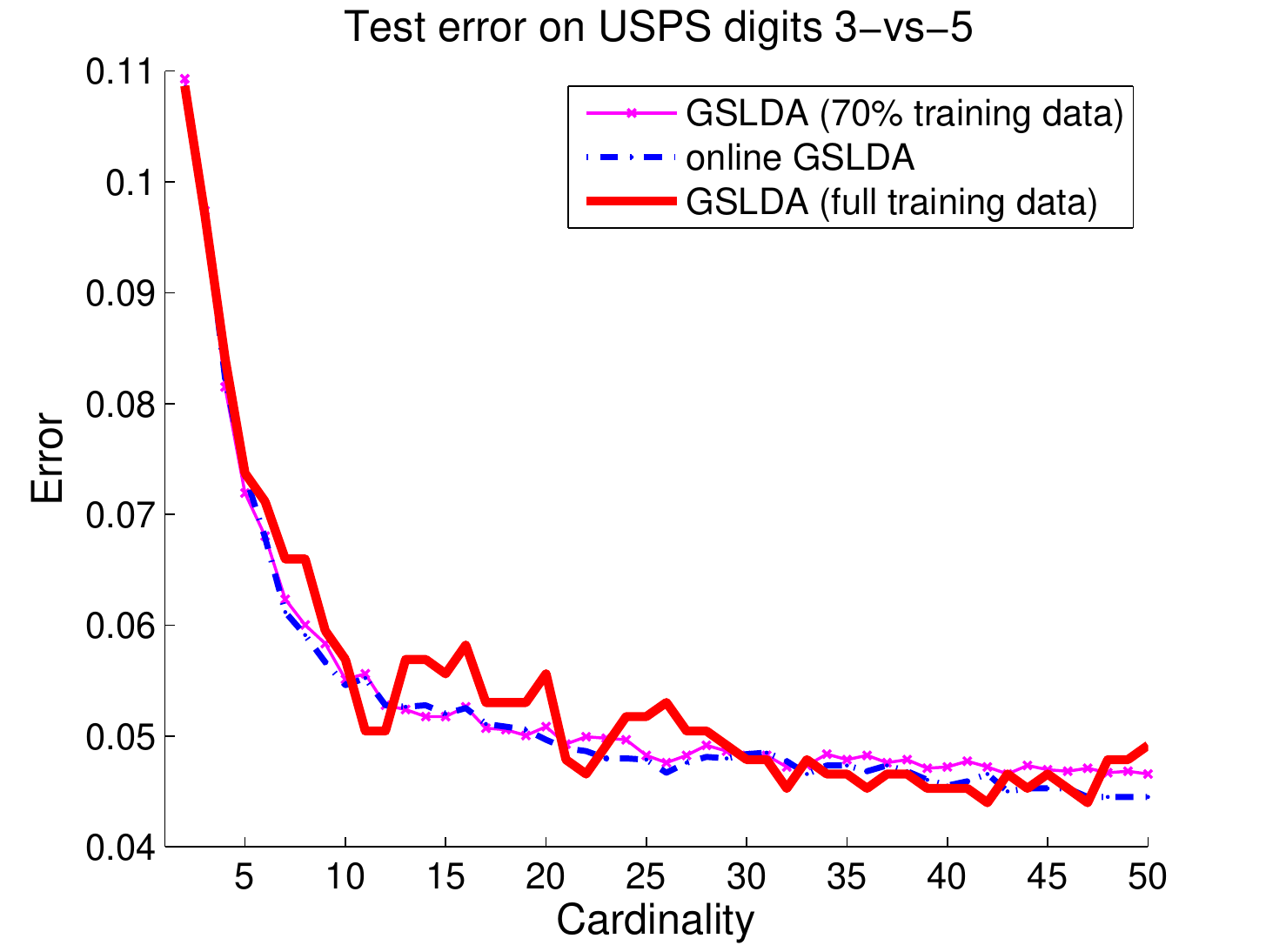}
        \label{fig:USPS1c}
      }
       \subfigure[]
      {
        \includegraphics[width=0.30\textwidth,clip]{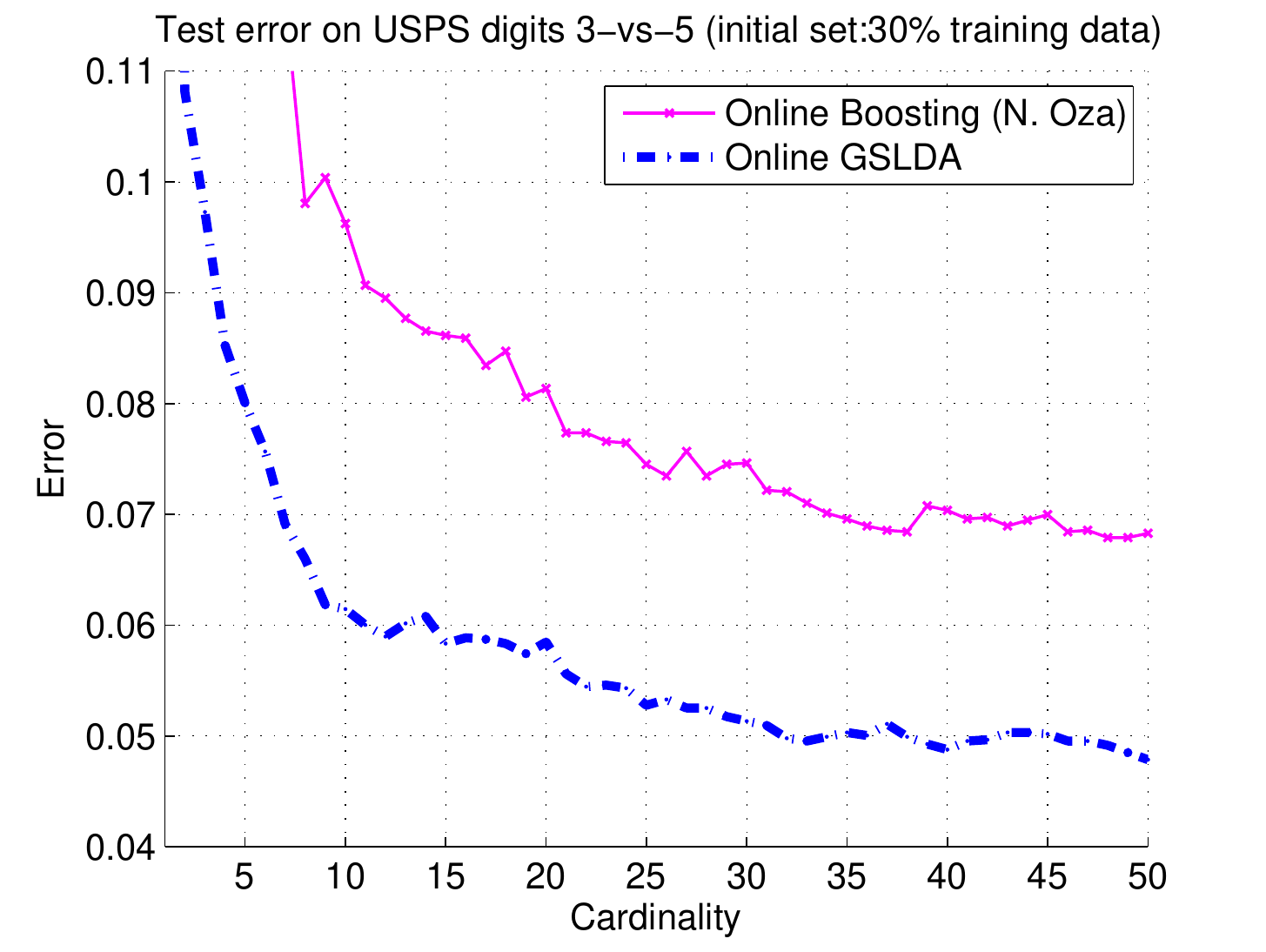}
        \label{fig:OGSLDA_OBOOST30}
      }
      \subfigure[]
      {
        \includegraphics[width=0.30\textwidth,clip]{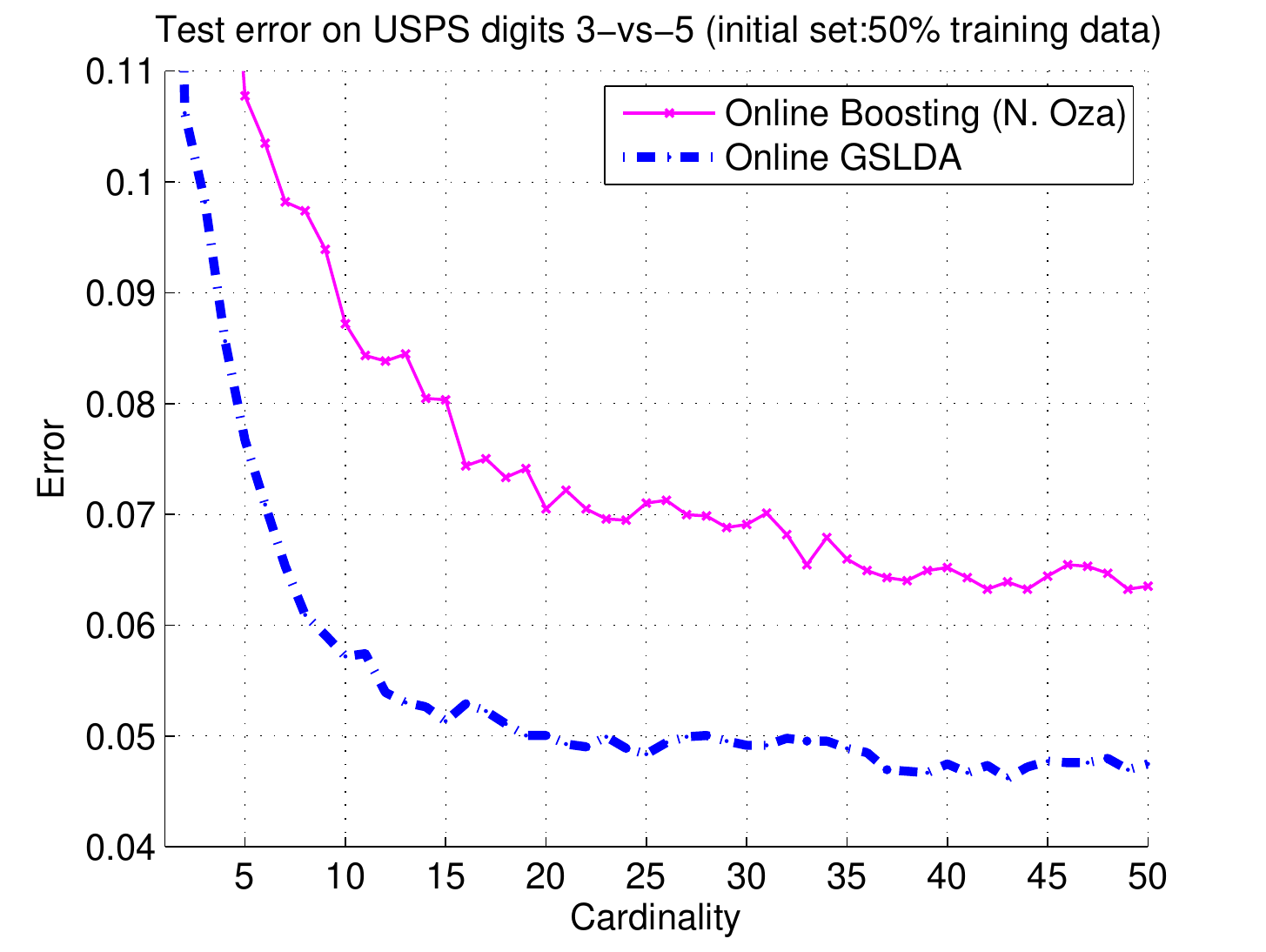}
        \label{fig:OGSLDA_OBOOST50}
      }
      \subfigure[]
      {
        \includegraphics[width=0.30\textwidth,clip]{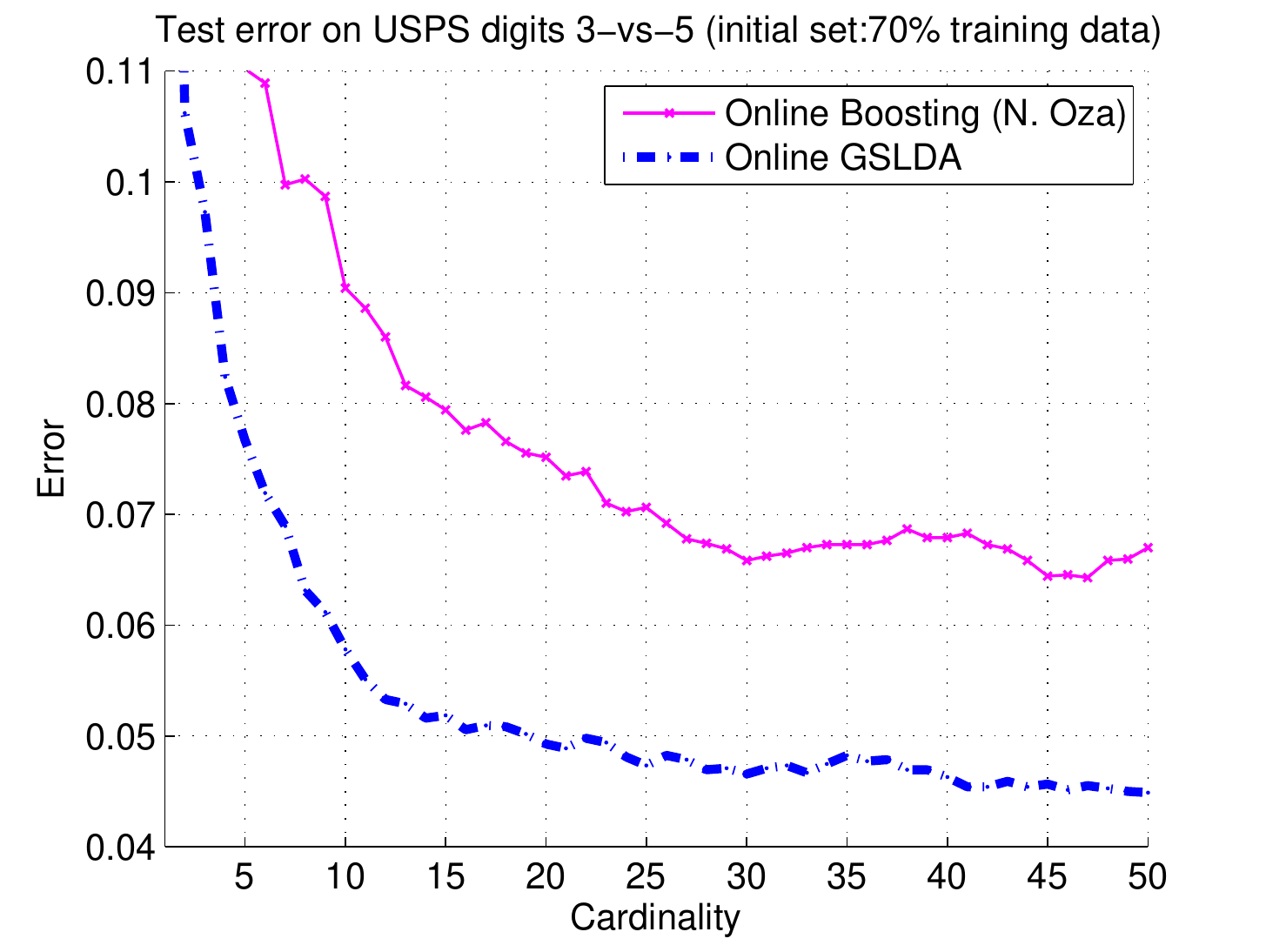}
        \label{fig:OGSLDA_OBOOST70}
      }
      \end{center}
      \caption{
        {\bf Top:} Classification error rates by offline GSLDA and online GSLDA
        on $16 \times 16$ pixels USPS digits data 
        sets \cite{Rasmussen2006Gaussian}.
        The number of initial training data for
        online GSLDA is 
        (a) $30\%$, (b) $50\%$, (c) $70\%$ of the available training data.
        All experiments, except batch GSLDA (trained with full training sets),
        are run $10$ times. The mean of the errors are plotted.
        {\bf Bottom:} Classification error rates by online GSLDA and 
        online boosting \cite{Oza2001Online}. The number of initial
        training data is (d) $30\%$, (e) $50\%$, (f) $70\%$ 
        of the available training data. All experiments are run $10$ times.
		}
	 \label{fig:USPS1}
  \end{figure*}

  \begin{figure*}[tbh!]
    \begin{center}
      \subfigure[]
      {
        \includegraphics[width=0.4\textwidth,clip]{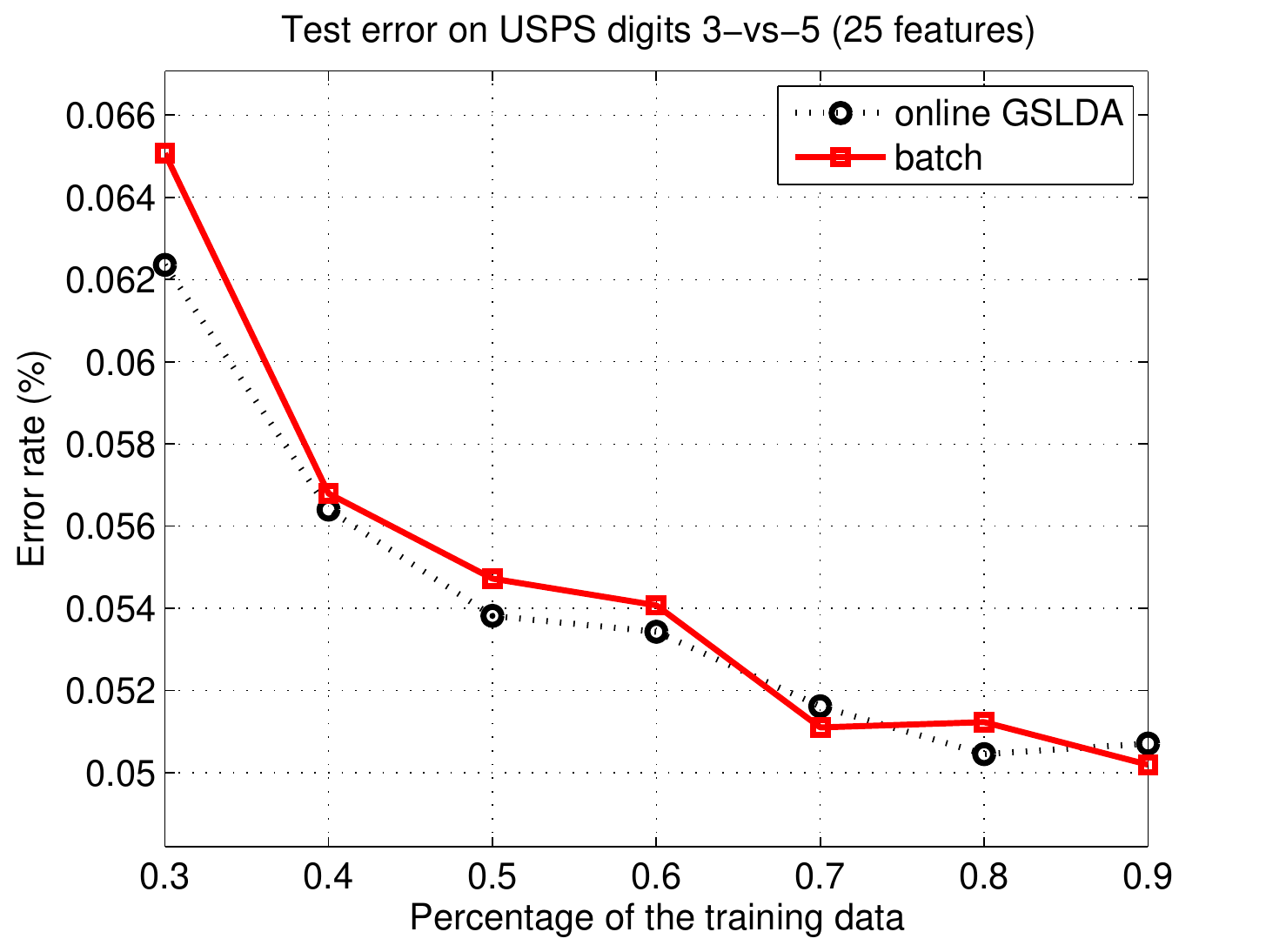}
        \label{fig:USPSerror25}
      }
      \subfigure[]
      {
        \includegraphics[width=0.4\textwidth,clip]{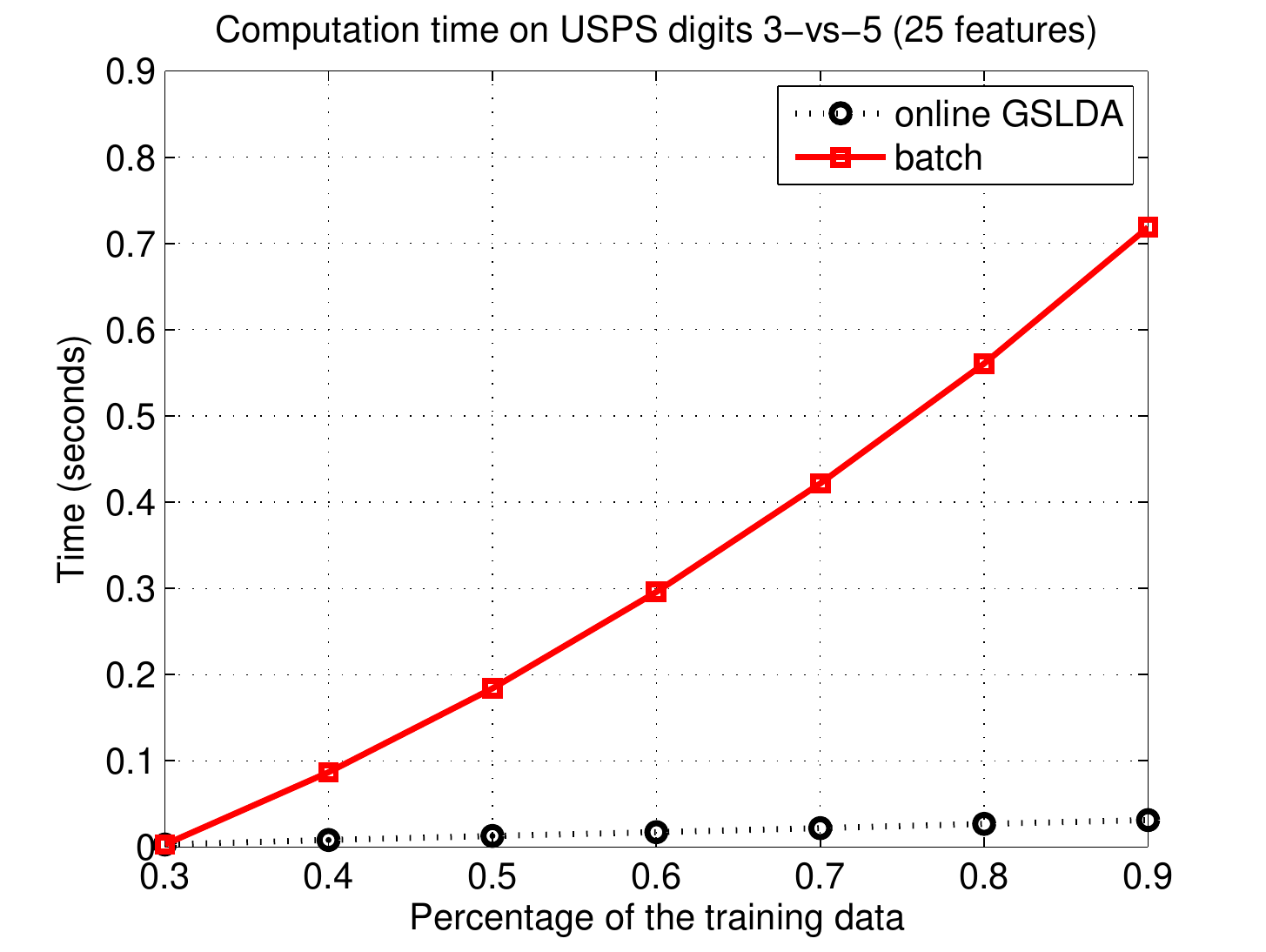}
        \label{fig:USPStime25}
      }
      \subfigure[]
      {
        \includegraphics[width=0.4\textwidth,clip]{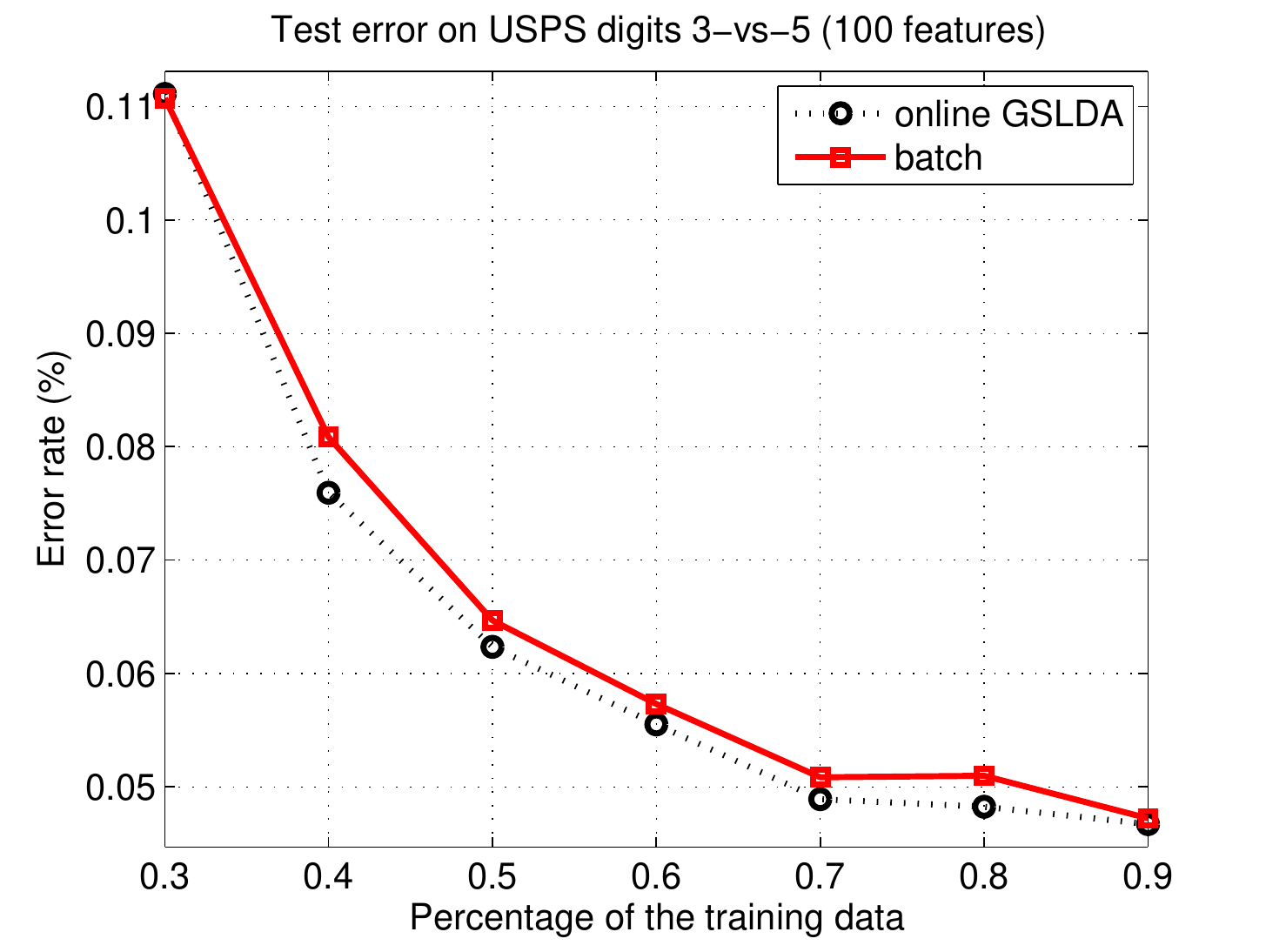}
        \label{fig:USPSerror100}
      }
      \subfigure[]
      {
        \includegraphics[width=0.4\textwidth,clip]{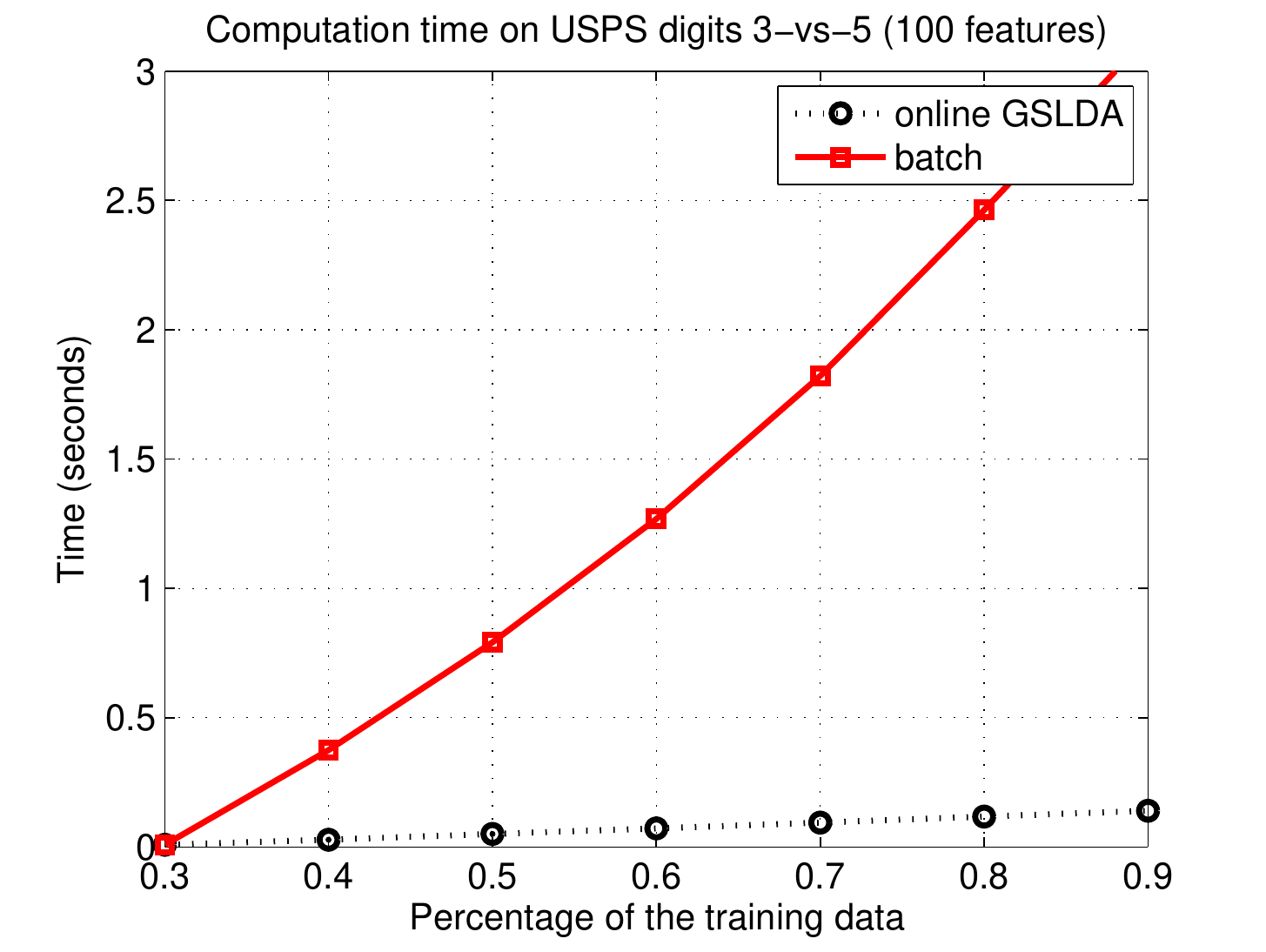}
        \label{fig:USPStime100}
      }
      \end{center}
      \caption{
        Comparison of classification error rate and 
        computation cost
        between online GSLDA and batch GSLDA on $16 \times 16$ pixels 
        USPS digits data sets \cite{Rasmussen2006Gaussian}. 
        We set the number of nonzero
        components of the feature coefficients ($\ell_0$ norm) to $25$ (a,b)
        and $100$ (c,d).
		}
	 \label{fig:USPS2}
  \end{figure*}

  We compare online GSLDA against batch GSLDA for classification of $16 \times 16$ pixels USPS
  digits `$3$' and `$5$'. The data set consists of $406$ training instances and $418$ test instances
  for the digit `$3$', $361$ training instances and $355$ test instances for digit `$5$'
  \cite{Rasmussen2006Gaussian}. We use the raw intensity value as the features. Hence, the total
  number of features is $256$. For batch learning, we applied greedy approach to sequentially select
  feature which yields maximal class separation (forward search). We then evaluate the performance
  of the classifier on the given test set and measure the error rate \cite{Moghaddam2007Fast}. For
  online learning, we randomly select $30/50/70$ percent training samples as the training set.
  Incremental updating is performed with the remaining training instances being inserted one at a
  time. We use decision stumps as the weak learners for both classifiers. All experiments, except
  batch GSLDA (trained with full training sets), are run $10$ times. The mean of the classification
  errors are plotted.

  Figs.~\ref{fig:USPS1a},~\ref{fig:USPS1b} and~\ref{fig:USPS1c} show the achieved classification
  error rates by batch GSLDA and online GSLDA. In the figures, the horizontal axis shows the
  $\ell_0$ norm of the feature coefficients, \ie, the number of weak classifiers,
  and the vertical axis indicates the classification error
  rate on test data. We observe a trend that the error rate decreases when we train with more
  training instances. It is important to point out that in this experiment the error rate of online
  GSLDA is quite close to that by batch GSLDA. We also train offline GSLDA classifiers with $30\%$,
  $50\%$ and $70\%$ training data. We observe an increase in error rates of GSLDA ($30\%$ training
  data) when the number of dimensions increase. This is not surprising since it is quite common for
  a classifier to overfit with large dimensions and small sample size.

  We compare the performance
  of online GSLDA with online boosting proposed in \cite{Oza2001Online}. For each weak classifier,
  we build a model by estimating the univariate normal distribution with weighted mean and variance
  for digits `$3$' and `$5$'. We update the weak classifier by incrementally updating the mean and
  variance using weighted version of \eqref{EQ:mtilde} and \eqref{EQ:sigmaTilde}. The results of
  online boosting are shown in Figs.~\ref{fig:OGSLDA_OBOOST30},~\ref{fig:OGSLDA_OBOOST50}
  and~\ref{fig:OGSLDA_OBOOST70}. The test error of online boosting decreases as the initial number
  of training samples increases. We observe that the performance of online boosting to be 
  remarkably worse
  than the performance of online GSLDA.

  Figs.~\ref{fig:USPSerror25} and~\ref{fig:USPSerror100} shows the achieved classification error
  rates by batch GSLDA and online GSLDA with $25$ and $100$ dimensions (features). In the figure,
  the horizontal axis shows the portion of training data instances and the vertical axis indicates
  the classification error rate. We observe a trend that the error rate decreases when more and more
  training data instances are involved, as expected.
  Online GSLDA not only performs well on this dataset but it
  is also very efficient. We give a comparison of the computation cost between batch GSLDA and
  incremental GSLDA in Figs.~\ref{fig:USPStime25} and \ref{fig:USPStime100}. As can be seen, the
  execution time of online GSLDA is significantly smaller than that of batch GSLDA as the number of
  training samples grows.

  \subsection{Frontal Face Detection}
  \label{sec:face}
	
  Due to its efficiency, Haar-like rectangle features \cite{Viola2004Robust} have become a popular choice as image features in the context of face detection. Similar to the work in \cite{Viola2004Robust}, the weak learning algorithm known as decision stumps and Haar-like rectangle features are used here due to their simplicity and efficiency. The following experiments compare offline GSLDA and online GSLDA learning algorithm.

    \subsubsection{Performances on Single-node Classifiers}
    \label{sec:onestage_gslda}

\begin{figure*}[tb!]
  \begin{center}
    \includegraphics[width=0.135\textwidth]{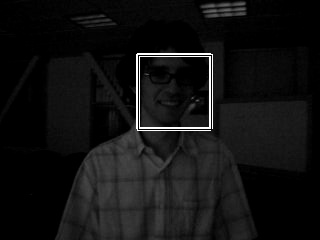}
    \includegraphics[width=0.135\textwidth]{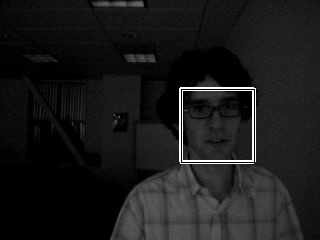}
    \includegraphics[width=0.135\textwidth]{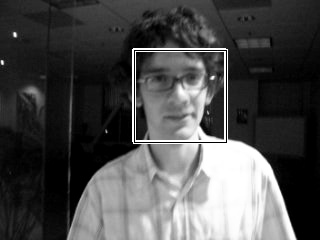}
    \includegraphics[width=0.135\textwidth]{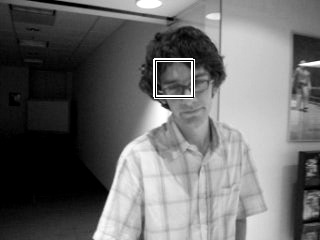}
    \includegraphics[width=0.135\textwidth]{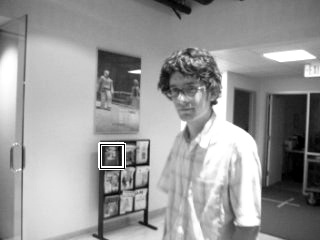}
    \includegraphics[width=0.135\textwidth]{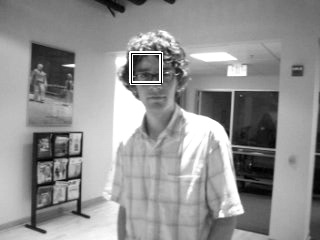}
    \includegraphics[width=0.135\textwidth]{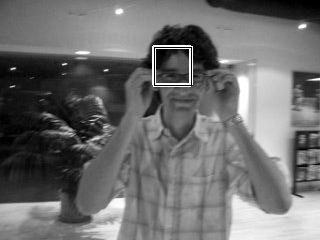}
    \includegraphics[width=0.135\textwidth]{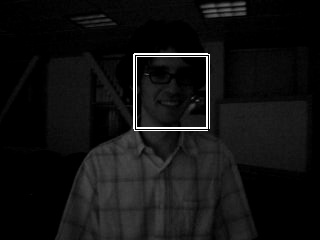}
    \includegraphics[width=0.135\textwidth]{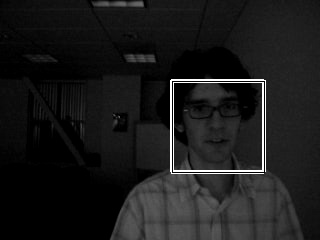}
    \includegraphics[width=0.135\textwidth]{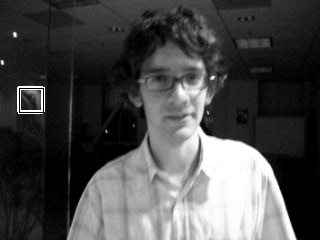}
    \includegraphics[width=0.135\textwidth]{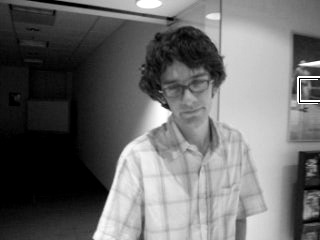}
    \includegraphics[width=0.135\textwidth]{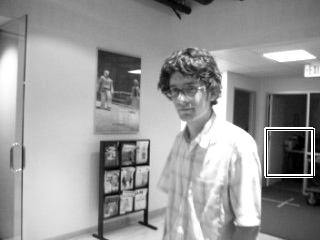}
    \includegraphics[width=0.135\textwidth]{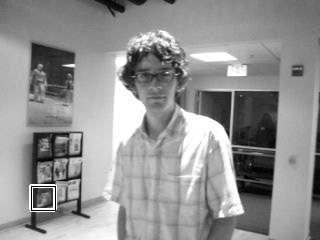}
    \includegraphics[width=0.135\textwidth]{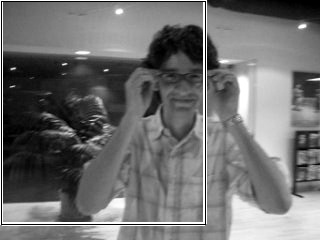}
    \includegraphics[width=0.135\textwidth]{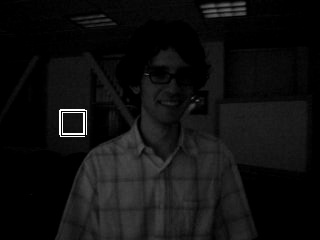}
    \includegraphics[width=0.135\textwidth]{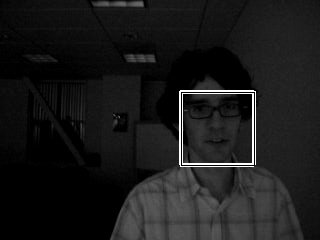}
    \includegraphics[width=0.135\textwidth]{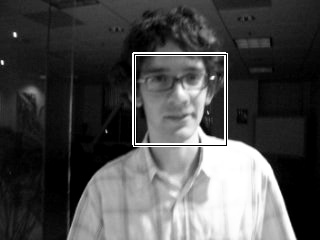}
    \includegraphics[width=0.135\textwidth]{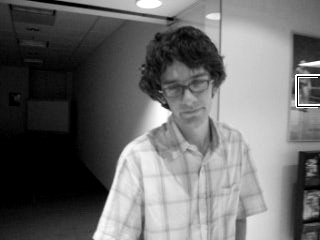}
    \includegraphics[width=0.135\textwidth]{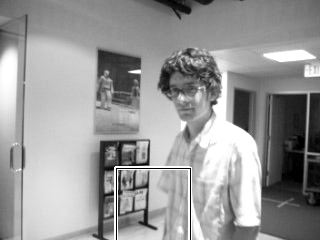}
    \includegraphics[width=0.135\textwidth]{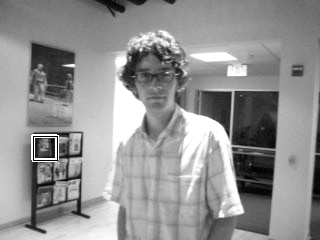}
    \includegraphics[width=0.135\textwidth]{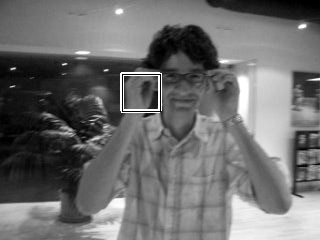}
    \includegraphics[width=0.135\textwidth]{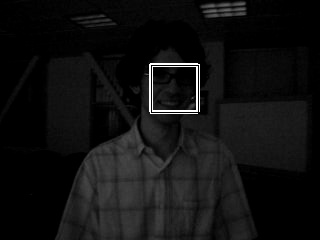}
    \includegraphics[width=0.135\textwidth]{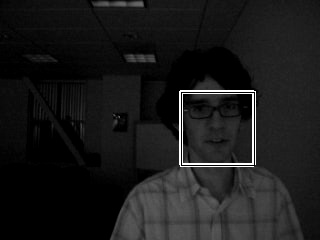}
    \includegraphics[width=0.135\textwidth]{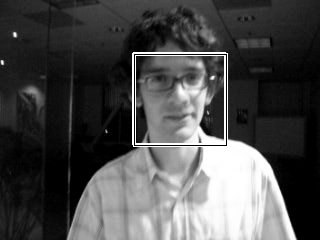}
    \includegraphics[width=0.135\textwidth]{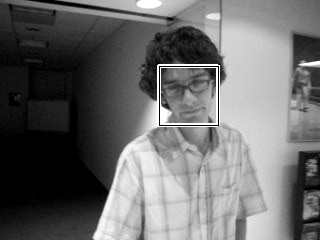}
    \includegraphics[width=0.135\textwidth]{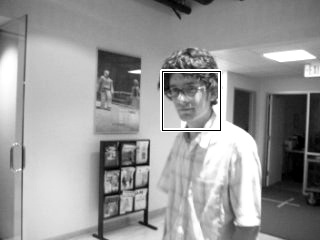}
    \includegraphics[width=0.135\textwidth]{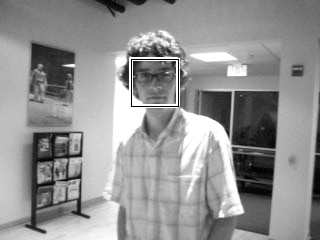}
    \includegraphics[width=0.135\textwidth]{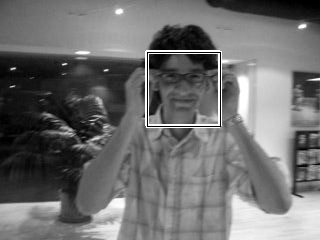}   

	\vspace{0.2cm}
    \includegraphics[width=0.135\textwidth]{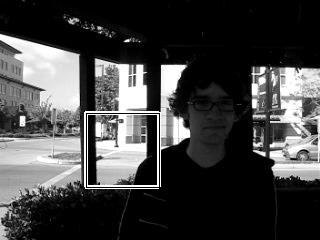}
    \includegraphics[width=0.135\textwidth]{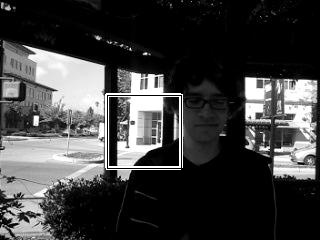}
    \includegraphics[width=0.135\textwidth]{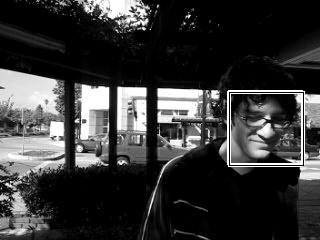}
    \includegraphics[width=0.135\textwidth]{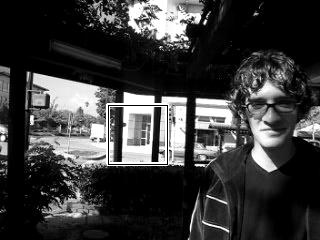}
    \includegraphics[width=0.135\textwidth]{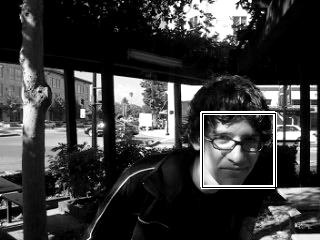}
    \includegraphics[width=0.135\textwidth]{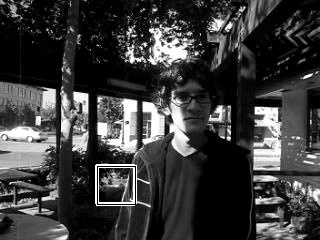}
    \includegraphics[width=0.135\textwidth]{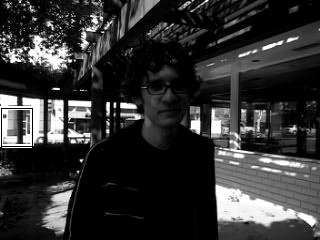}    
    \includegraphics[width=0.135\textwidth]{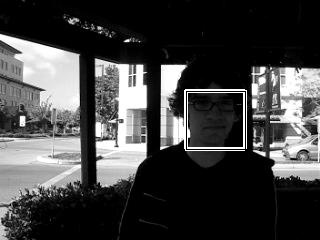}
    \includegraphics[width=0.135\textwidth]{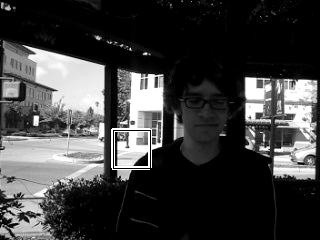}
    \includegraphics[width=0.135\textwidth]{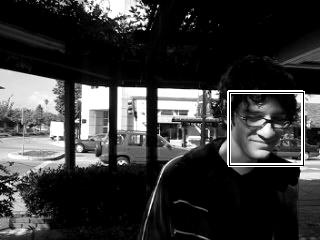}
    \includegraphics[width=0.135\textwidth]{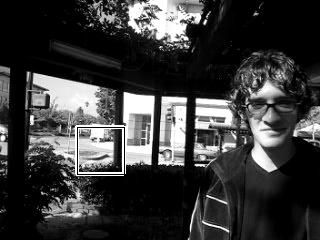}
    \includegraphics[width=0.135\textwidth]{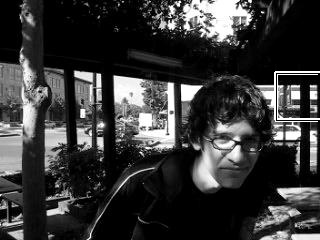}
    \includegraphics[width=0.135\textwidth]{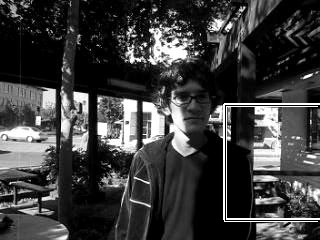}
    \includegraphics[width=0.135\textwidth]{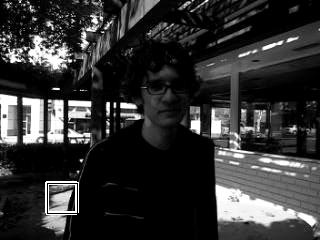}    
    \includegraphics[width=0.135\textwidth]{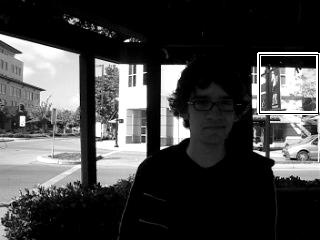}
    \includegraphics[width=0.135\textwidth]{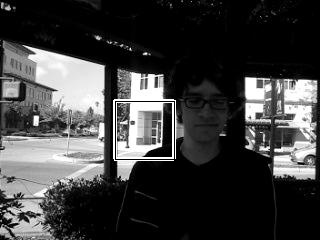}
    \includegraphics[width=0.135\textwidth]{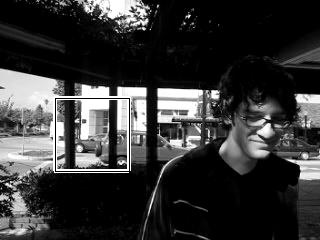}
    \includegraphics[width=0.135\textwidth]{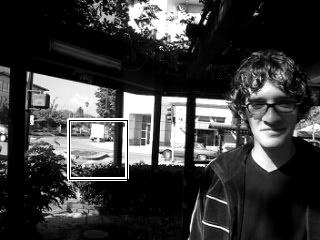}
    \includegraphics[width=0.135\textwidth]{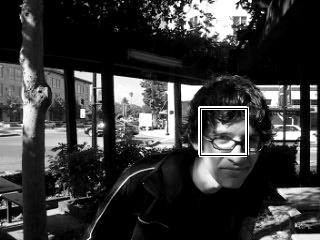}
    \includegraphics[width=0.135\textwidth]{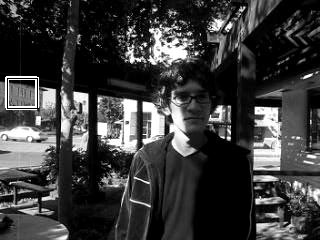}
    \includegraphics[width=0.135\textwidth]{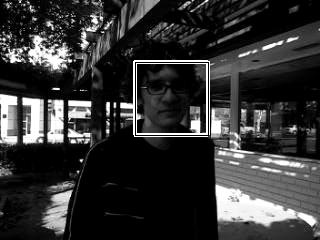}    
    \includegraphics[width=0.135\textwidth]{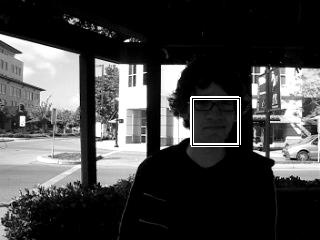}
    \includegraphics[width=0.135\textwidth]{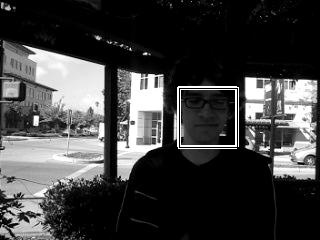}
    \includegraphics[width=0.135\textwidth]{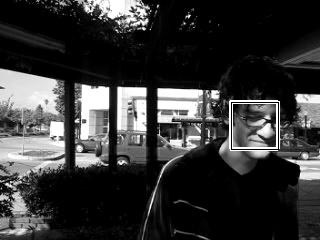}
    \includegraphics[width=0.135\textwidth]{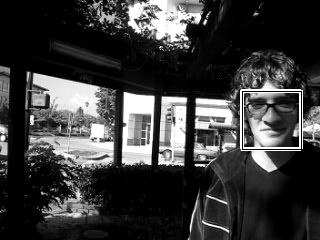}
    \includegraphics[width=0.135\textwidth]{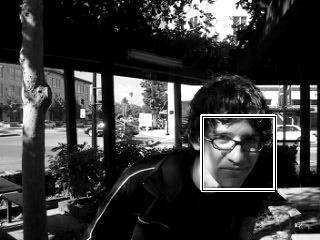}
    \includegraphics[width=0.135\textwidth]{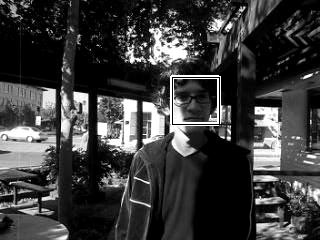}
    \includegraphics[width=0.135\textwidth]{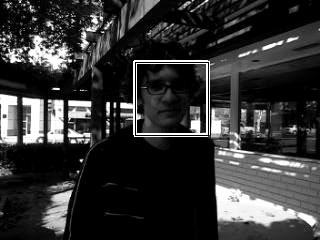}  
  \end{center} 	
  \caption{A comparison of offline AdaBoost based frontal face detector
  \cite{Viola2004Robust} ({\bf Top row}), AsymBoost based face detector
  \cite{Viola2002Fast} ({\bf Second row}), GSLDA based face detector 
  \cite{Paisitkriangkrai2009CVPR} ({\bf Third row}) and our proposed
  OGSLDA face detector ({\bf Last row}). All detectors are
  trained initially with $1,000$ faces and $10,000$ non-faces. Online GSLDA
  is incrementally updated with patches classified as faces from the previous
  video frames. The first video ({\em david indoor}) contains $761$ frames of
  a person moving from a dark to a bright area undergoing large lighting
  and pose changes 
  (frames $150$, $250$, $350$, $409$, $450$, $494$ and $592$). 
  The second video (\emph{trellis}) contains $501$ frames of
  a person moving underneath a trellis with large illumination change
  (frames $50$, $85$, $182$, $231$, $287$, $386$ and $457$).}
  \label{fig:davidFrames}
 \end{figure*}

\begin{figure}[b!]
            \begin{center}
\includegraphics[width=0.03\textwidth]{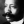}
\includegraphics[width=0.03\textwidth]{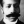}
\includegraphics[width=0.03\textwidth]{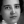}
\includegraphics[width=0.03\textwidth]{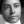}
\includegraphics[width=0.03\textwidth]{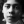}
\includegraphics[width=0.03\textwidth]{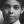}
\includegraphics[width=0.03\textwidth]{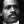}
\includegraphics[width=0.03\textwidth]{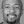}
\includegraphics[width=0.03\textwidth]{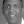}
\includegraphics[width=0.03\textwidth]{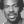}
\includegraphics[width=0.03\textwidth]{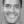}
\includegraphics[width=0.03\textwidth]{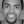}
\includegraphics[width=0.03\textwidth]{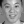}
\includegraphics[width=0.03\textwidth]{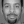}
\includegraphics[width=0.03\textwidth]{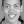}
\includegraphics[width=0.03\textwidth]{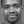}
\includegraphics[width=0.03\textwidth]{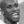}
\includegraphics[width=0.03\textwidth]{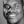}
\includegraphics[width=0.03\textwidth]{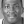}
\includegraphics[width=0.03\textwidth]{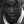}
\includegraphics[width=0.03\textwidth]{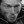}
\includegraphics[width=0.03\textwidth]{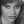}
\includegraphics[width=0.03\textwidth]{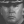}
\includegraphics[width=0.03\textwidth]{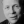}
\includegraphics[width=0.03\textwidth]{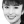}
\includegraphics[width=0.03\textwidth]{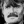}
\includegraphics[width=0.03\textwidth]{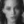}
\includegraphics[width=0.03\textwidth]{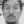}
\includegraphics[width=0.03\textwidth]{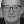}
\includegraphics[width=0.03\textwidth]{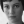}
\includegraphics[width=0.03\textwidth]{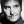}
\includegraphics[width=0.03\textwidth]{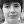}
\includegraphics[width=0.03\textwidth]{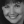}
\includegraphics[width=0.03\textwidth]{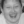}
\includegraphics[width=0.03\textwidth]{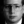}
\includegraphics[width=0.03\textwidth]{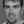}
\includegraphics[width=0.03\textwidth]{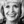}
\includegraphics[width=0.03\textwidth]{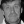}
\includegraphics[width=0.03\textwidth]{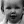}
\includegraphics[width=0.03\textwidth]{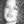}
\includegraphics[width=0.03\textwidth]{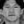}
\includegraphics[width=0.03\textwidth]{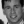}
\includegraphics[width=0.03\textwidth]{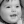}
\includegraphics[width=0.03\textwidth]{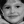}
\includegraphics[width=0.03\textwidth]{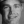}
\includegraphics[width=0.03\textwidth]{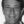}
\includegraphics[width=0.03\textwidth]{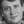}
\includegraphics[width=0.03\textwidth]{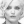}
\includegraphics[width=0.03\textwidth]{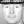}
\includegraphics[width=0.03\textwidth]{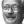}
\includegraphics[width=0.03\textwidth]{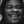}
\includegraphics[width=0.03\textwidth]{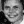}
\includegraphics[width=0.03\textwidth]{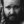}
\includegraphics[width=0.03\textwidth]{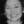}
\includegraphics[width=0.03\textwidth]{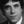}
\includegraphics[width=0.03\textwidth]{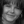}
\includegraphics[width=0.03\textwidth]{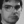}
\includegraphics[width=0.03\textwidth]{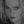}
\includegraphics[width=0.03\textwidth]{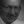}
\includegraphics[width=0.03\textwidth]{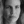}
\includegraphics[width=0.03\textwidth]{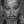}
\includegraphics[width=0.03\textwidth]{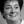}
\includegraphics[width=0.03\textwidth]{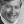}
\includegraphics[width=0.03\textwidth]{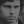}
\includegraphics[width=0.03\textwidth]{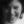}
\end{center}
            \caption{A random sample of face  images for training.}
            \label{FIG:training_faces}
        \end{figure}

    We conduct two experiments in this section. The first experiment compares single strong
    classifier learned using AdaBoost \cite{Viola2004Robust}, AsymBoost \cite{Viola2002Fast},
    offline GSLDA \cite{Paisitkriangkrai2009CVPR} and our proposed online GSLDA algorithms. The
    datasets consist of $1,000$ mirrored face examples (Fig.~\ref{FIG:training_faces}) and $10,000$
    bootstrapped non-face examples. The face were cropped and rescaled to images of size $24 \times
    24$ pixels. For non-face examples, we initially select $1,000$ random non-face patches from
    non-face images. The other $9,000$ non-face patches are added to the initial pool of training
    data by bootstrapping\footnote{We incrementally construct new non-face samples using a trained
    classifier of \cite{Viola2004Robust}.}.

    We train three offline face detectors using AdaBoost, AsymBoost and GSLDA. Each classifier
    consists of 200 weak classifiers. The classifiers are tested on a challenged face videos, David
    Ross indoor data set and trellis data set\footnote{\url{http://www.cs.toronto.edu/~dross/ivt/}},
    which are publicly available on the internet. Both videos contain large lighting variation, cast
    shadows, unknown camera motion, and tilted face with in-plane and out-of-plane rotation. The
    first video contains $761$ frames of a person moving from a dark to a bright area. Since, the
    first few video frames has very low contrast (almost impossible to see faces), we ignore the
    first $100$ frames. The second video contains $501$ frames of a person moving underneath a
    trellis with large illumination change and cast shadows.

    In this experiment, we use the scanning window technique to locate faces. We set the scaling
    factor to $1.2$ and window shifting step to $1$. The patch with highest classification score is
    classified as faces. In other words, there is only one selected face in each frame. The criteria
    similar to the one used in PASCAL VOC Challenge \cite{Pascal07} is adopted here. Detections are
    considered true or false positives based on the area of overlap with ground truth bounding
    boxes. To be considered a correct detection, the area of overlap between the predicted bounding
    box, $B_p$, and ground truth bounding box, $B_{gt}$, must exceed 50\% by the formula:
\[
  \frac{area(B_p \cap B_{gt})}{area(B_p \cup B_{gt})} > 50\%.
\]
For online GSLDA, the predicted faces in the previous frames are used to update the GSLDA model.
Note that the updated patches could contain both true positives (faces) and false positives
(misclassified non-faces). After the update process, the classifier predicts a single patch with
highest classification score in the next frame as the face patch. This learning technique is similar
to semi-supervised learning where the classifier makes use of the unlabeled data in conjunction with
a small amount of labeled data. Note that unlike the work in \cite{Grabner2006Online} where both
positive and negative patches are used to incrementally update their model, we only make use of
positive patches. 

Table~\ref{tab:tabross} compares the four face detectors in terms of their performance. We observe
that the performance of AdaBoost face detector is the worst. This is not surprising since the
distributions of training data are highly skewed ($1,000$ faces and $10,000$ non-faces). Viola and
Jones also pointed out this limitation in \cite{Viola2002Fast}. Face detectors trained using
AsymBoost and GSLDA perform quite similar on the first video. The results are consistent with the
ones reported in \cite{Paisitkriangkrai2009CVPR}. Our results show that online GSLDA performs best.
Based on our observations, incrementally updating GSLDA model improves the detection results
significantly at small increase in computation time. Fig.~\ref{fig:davidFrames} compares the
empirical results between offline GSLDA and our proposed online GSLDA.

\begin{table}[tb!]
  \caption{Performance on four different frontal face detectors on david indoor and trellis video}
  \begin{center}
    \begin{tabular}{l|c|c}
    \hline
      & \multicolumn{2}{c}{ detection rate} \\
    \cline{2-3}
    $ $ & indoor sequence & trellis sequence \\
    \hline
    \hline
      AdaBoost \cite{Viola2004Robust} &	$57.8\%$ & $35.3\%$ \\
      AsymBoost \cite{Viola2002Fast}  & $68.7\%$ & $37.5\%$ \\
      GSLDA \cite{Paisitkriangkrai2009CVPR}	& $70.3\%$ & $48.5\%$ \\
      Our proposed OGSLDA                   & $83.1\%$ & $62.1\%$ \\
     \hline
    \end{tabular}
  \end{center}
  \label{tab:tabross}
\end{table}

\begin{figure}[t]
  \begin{center}
    \includegraphics[width=0.4\textwidth,clip]{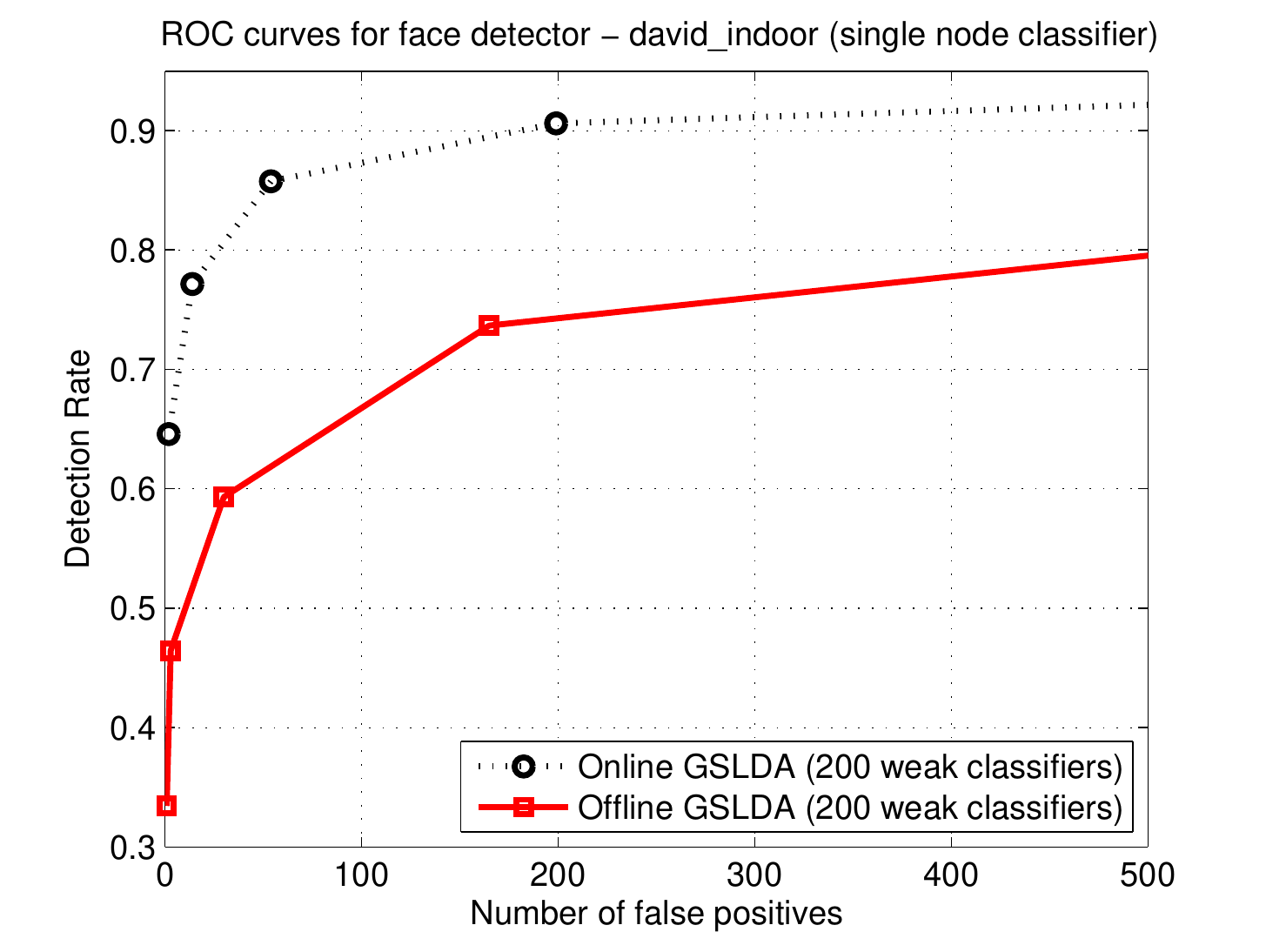}
    \includegraphics[width=0.4\textwidth,clip]{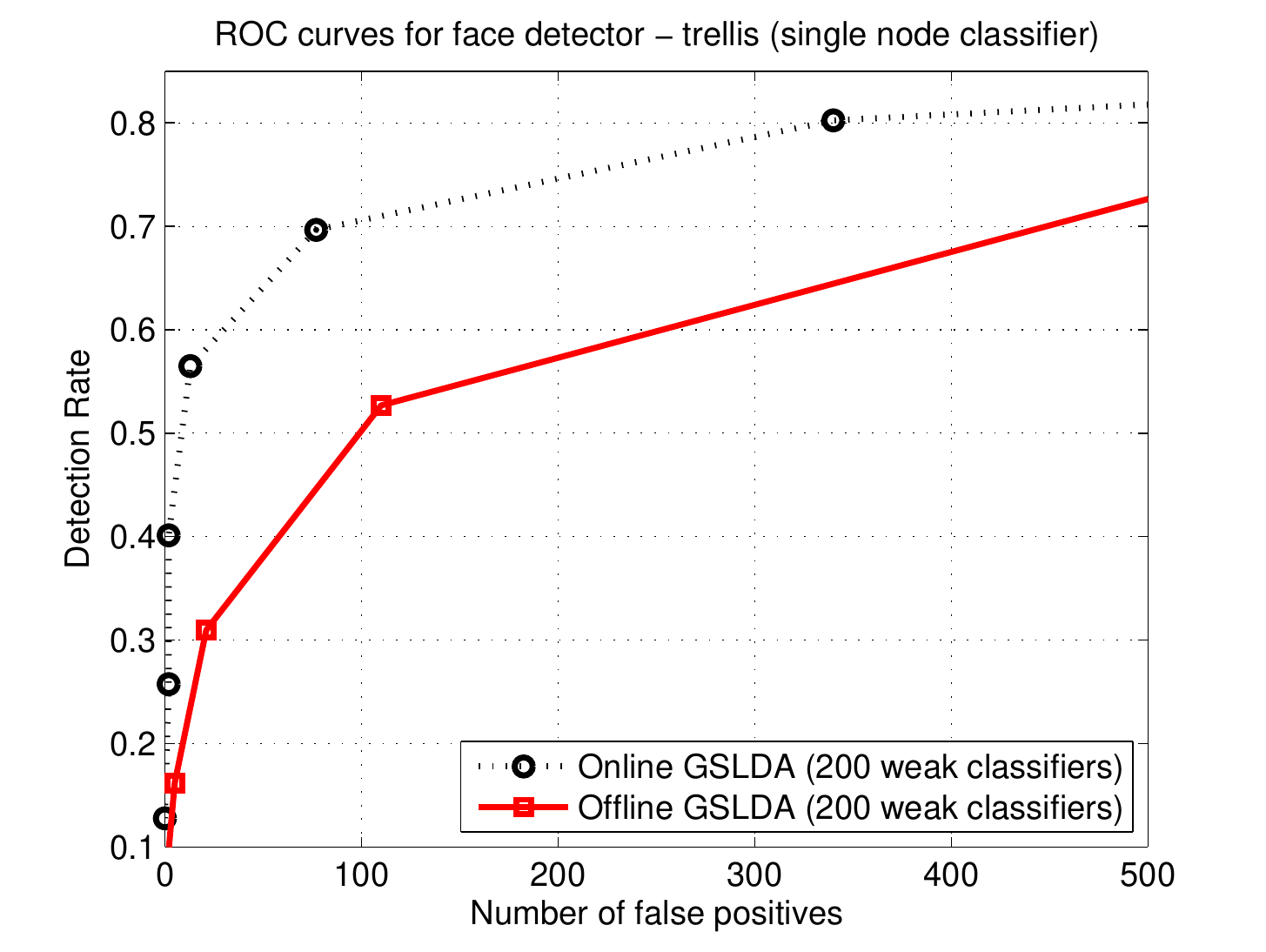}
  \end{center}
  \caption{
    Comparison of ROC curves between offline and online GSLDA
    on David Ross indoor data set (top) and trellis data set
    (bottom).
  }
  \label{fig:davidROC}
\end{figure}

    Finally, we compare the Receiver Operating Characteristic (ROC) curves between the offline GSLDA
    model ($1,000$ faces and $10,000$ non-faces) and the online GSLDA model (initially trained with
    $1,000$ faces and $10,000$ non-faces + updated with $661$ patches classified as faces). In this
    experiment, we set the scaling factor to $1.2$ and window stepping size to $1$. The techniques
    used for merging overlapping windows are similar to \cite{Viola2004Robust}. Detections are
    considered true or false positives based on the area overlap with ground truth bounding boxes.
    We shift the classifier threshold and plot the ROC curves (Fig.~\ref{fig:davidROC}). Clearly,
    updating the trained model with relevant training data increases the overall performance of the
    classifiers.

\begin{table}[t]
    \caption{The size of training and test sets used on the single node classifier.}
	\begin{center}
	\begin{tabular}{l|ccc}
	\hline
	$  $\#$ $ & data splits  &  faces$/$split & non-faces$/$split\\
		\hline
        \hline
			Train   &  $3$ & $2000$ & $2000$  \\
			Test    &  $2$ & $2000$ & $2000$  \\
		\hline
	\end{tabular}
	\end{center}
\label{tab:1}
\end{table}

    In the next experiment, we compare the performance of single strong classifiers learned using
    offline GSLDA and online GSLDA algorithms on frontal faces database. The database consists of
    $10,000$ mirrored faces. The faces were cropped and rescaled to images of size $24 \times 24$
    pixels. For non-face examples, we randomly selected 10,000 random non-face patches from non-face
    images obtained from the internet. The collected patches are split into three training sets and
    two test sets. Each set contains 2,000 face examples and 2,000 non-face examples
    (Table~\ref{tab:1}). For each experiment, three different classifiers are generated, each by
    selecting two out of three training sets and the remaining training set for validation.

\begin{figure*}[tbh!]
    \begin{center}
      \subfigure[]
      {
        \includegraphics[width=0.30\textwidth,clip]{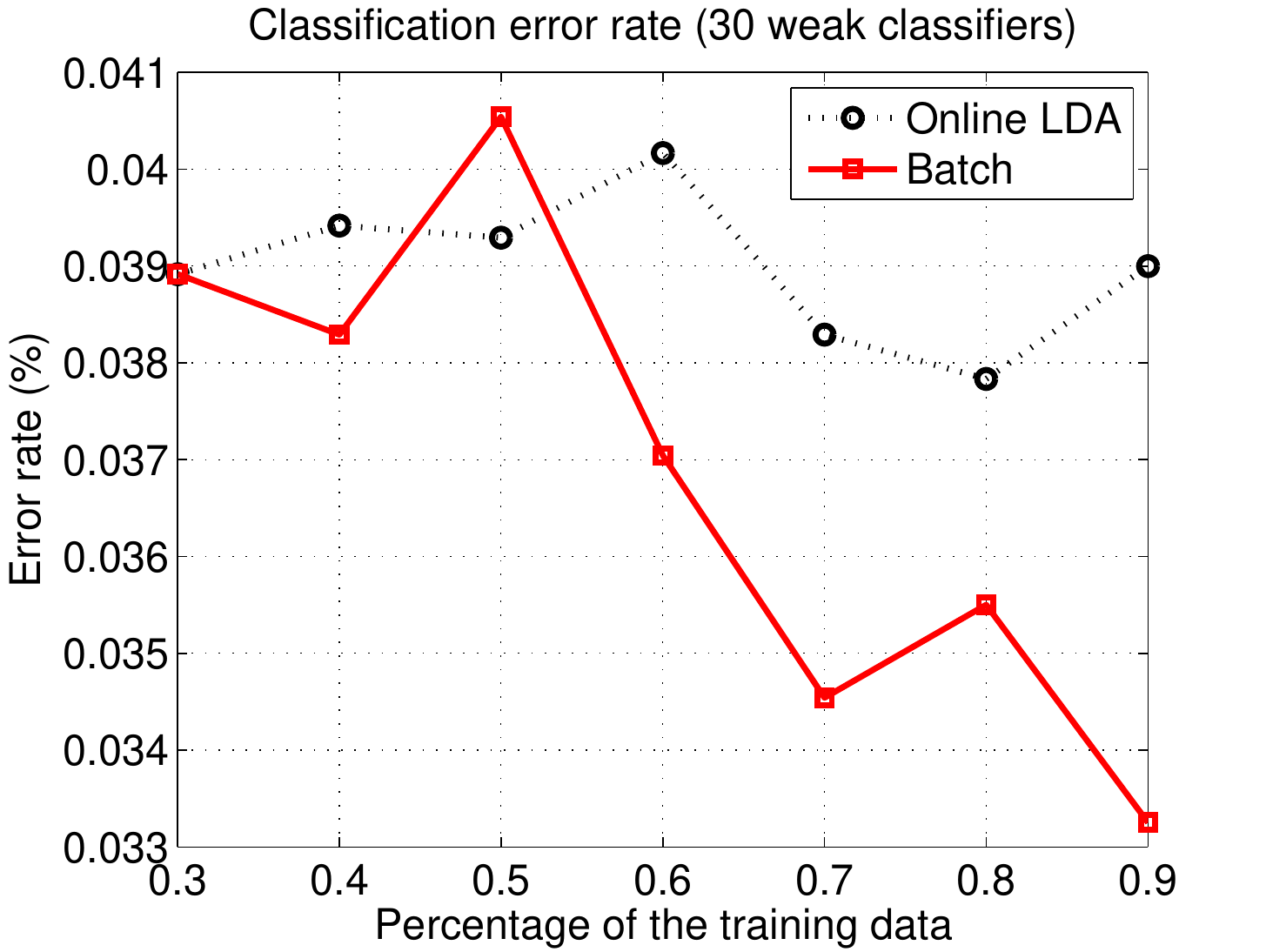}
      }
      \subfigure[]
      {
        \includegraphics[width=0.30\textwidth,clip]{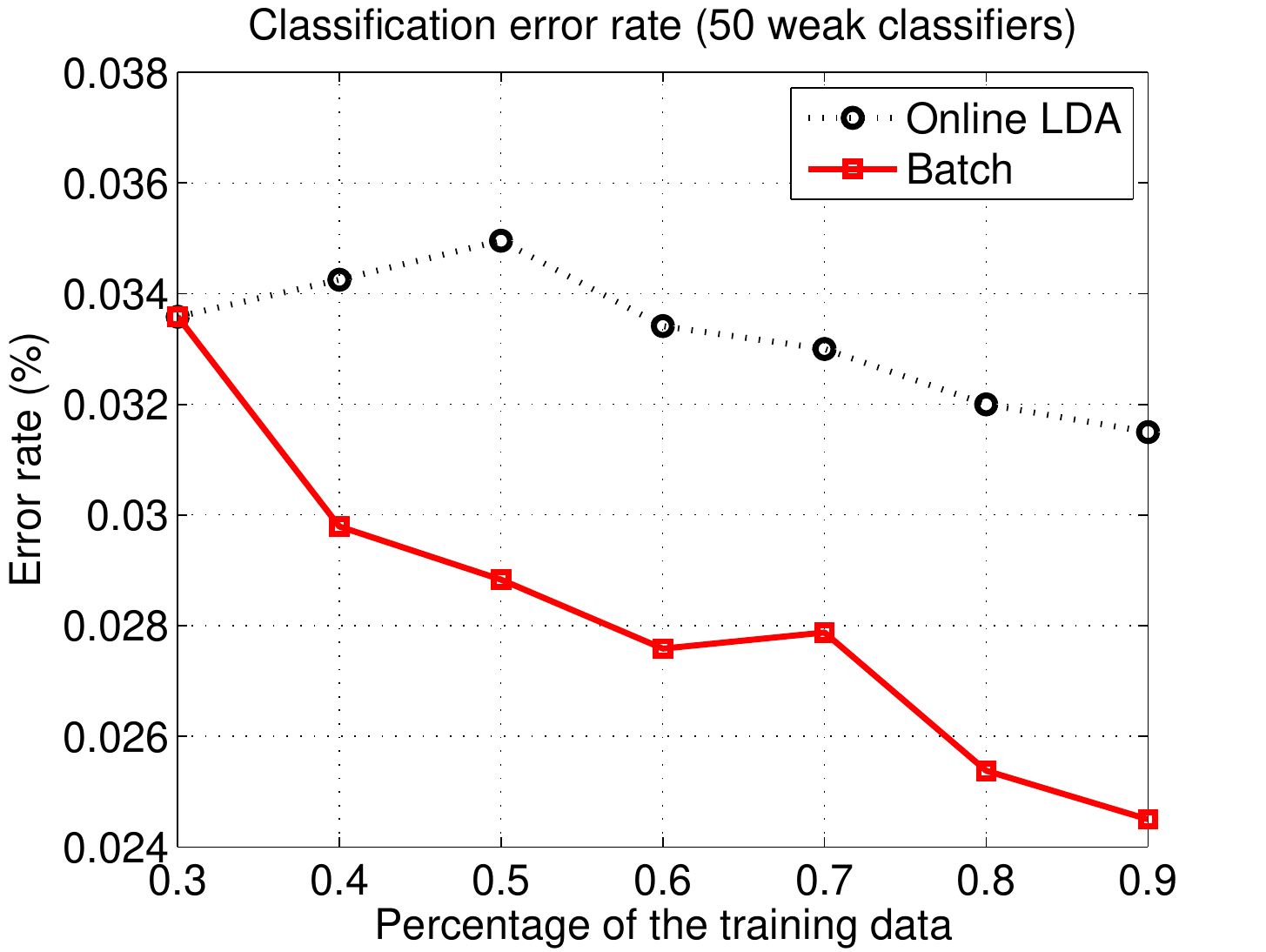}
      }
      \subfigure[]
      {
        \includegraphics[width=0.30\textwidth,clip]{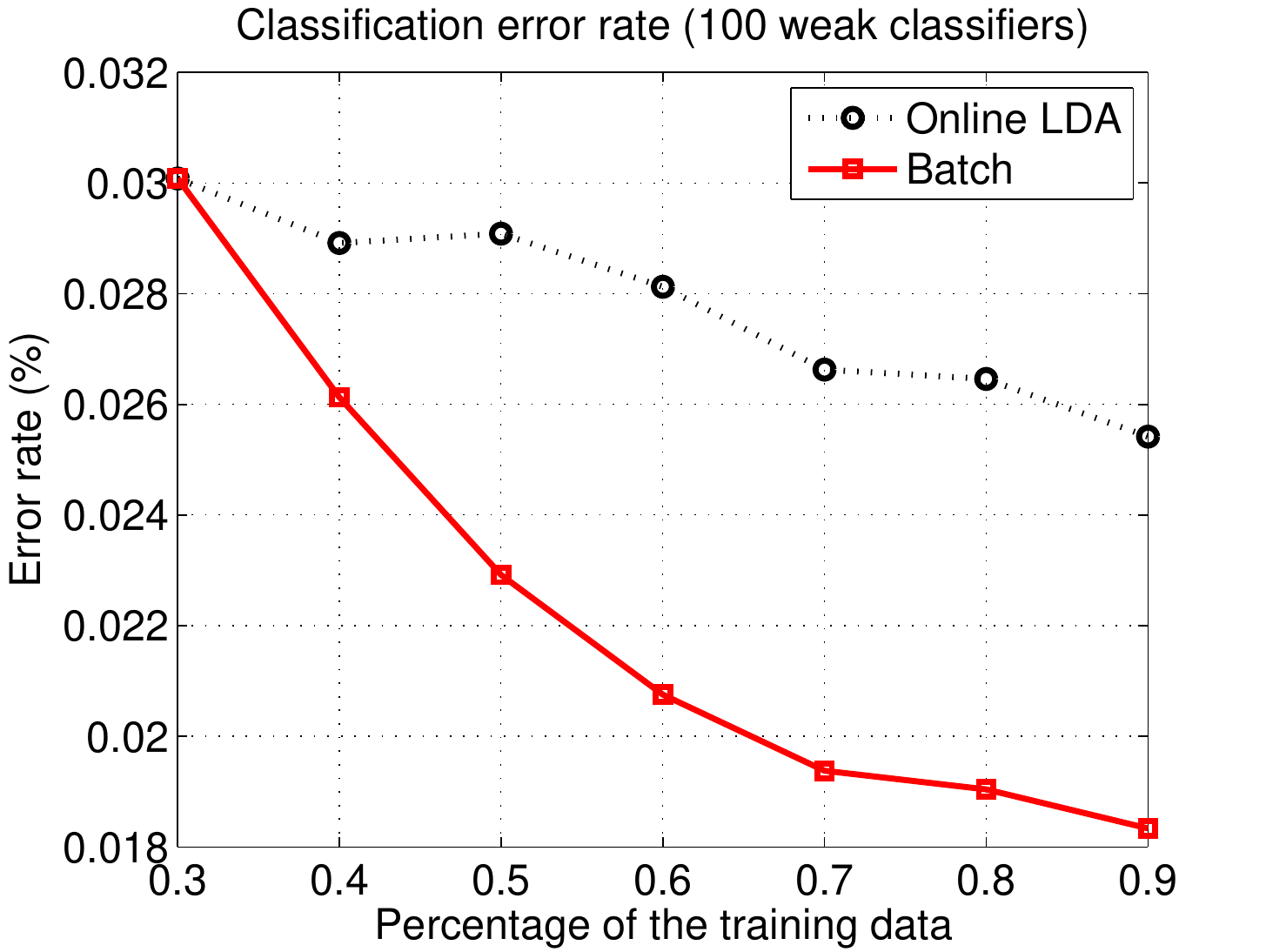}
      }
      \end{center}
      \caption{
        Comparison of classification error rates between batch GSLDA
        and online GSLDA. The number of weak learners (decision stumps on 
        Haar-like features) in each experiment is
        (a) $30$, (b) $50$, (c) $100$. 
        The error of both classifiers drops as the number of 
        training samples increases.}
	 \label{fig:single_face}
  \end{figure*}

    In this experiment, we train $30$, $50$ and $100$ weak learners of Haar-like features. The
    performance is measured by the test error rate. The results are shown in
    Fig.~\ref{fig:single_face}. The following observations can be made from these curves. The error
    of both classifiers drops as the number of training samples increases. The error rate of batch
    GSLDA drops at a slightly faster rate than online GSLDA. This is not surprising. For batch
    learning, the previous set of training samples along with a new sample are used to update the
    decision stumps every time a new sample is inserted. For each update, GSLDA algorithm throws
    away previously selected weak classifiers and reselects the new $30$, $50$ and $100$ weak
    classifiers. As a result, the training process is time consuming and requires a large amount of
    storage. In contrast, online GSLDA relies on the initial trained decision stumps. The new
    instance does not update the trained decision stumps but the between-class and within-class
    scatter matrices. The process is suboptimal compared to batch GSLDA. However, the slight
    increase in performance of batch GSLDA over online GSLDA ($0.7\%$ drop in test error rate for
    $100$ weak classifiers) comes at a much higher storage cost and significantly higher computation
    time.

	\subsubsection{Performances on Cascades of Strong Classifiers}
	\label{sec:cascade_gslda}

    \begin{figure*}[tbh!]
      \begin{center}
        \subfigure[]
        {
          \includegraphics[width=0.30\textwidth,clip]{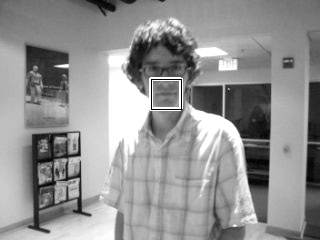}
          \label{fig:cascade_face_a}
        }
        \subfigure[]
        {
          \includegraphics[width=0.30\textwidth,clip]{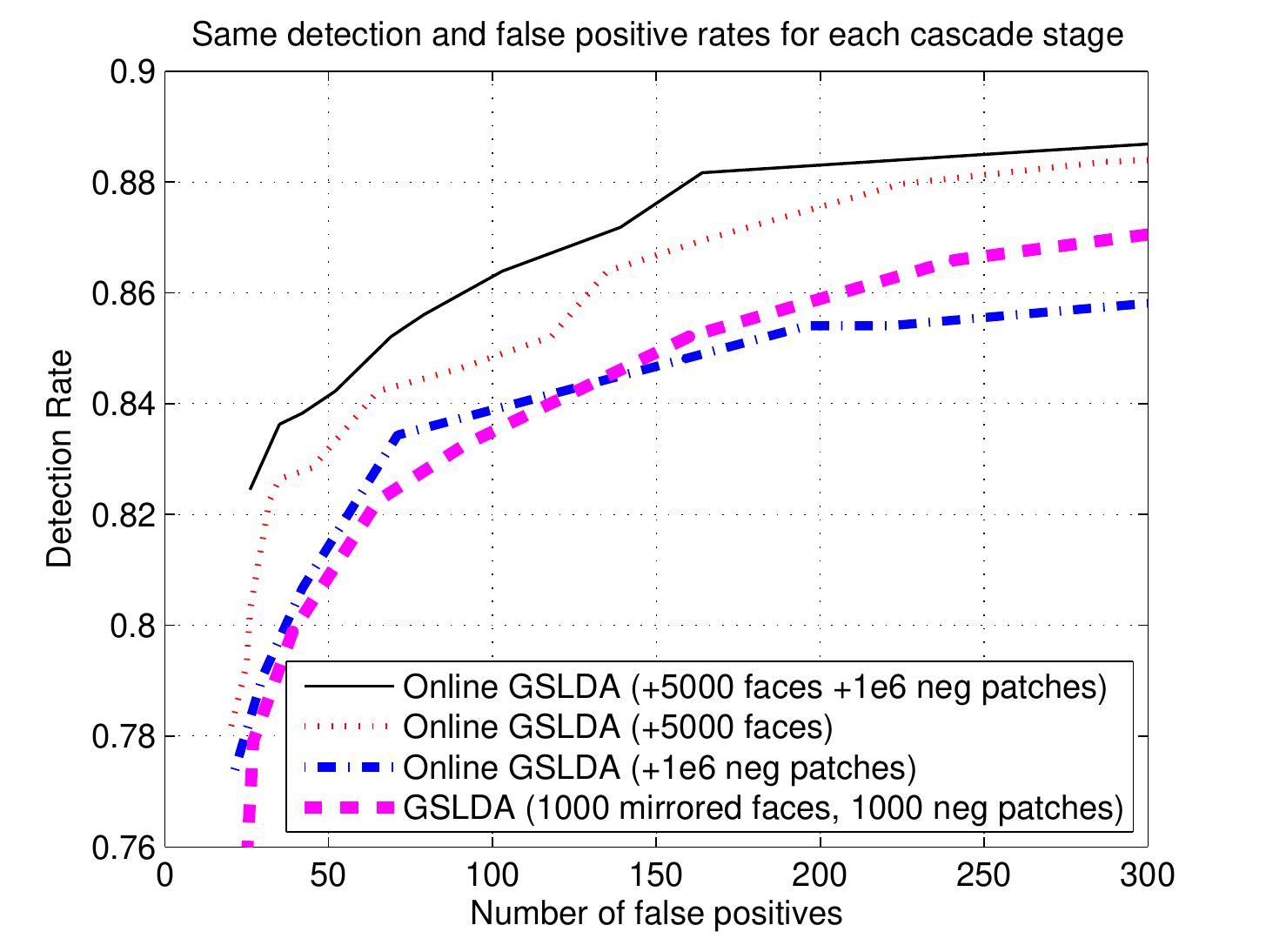}
          \label{fig:cascade_face_b}
        }
        \subfigure[]
        {
          \includegraphics[width=0.30\textwidth,clip]{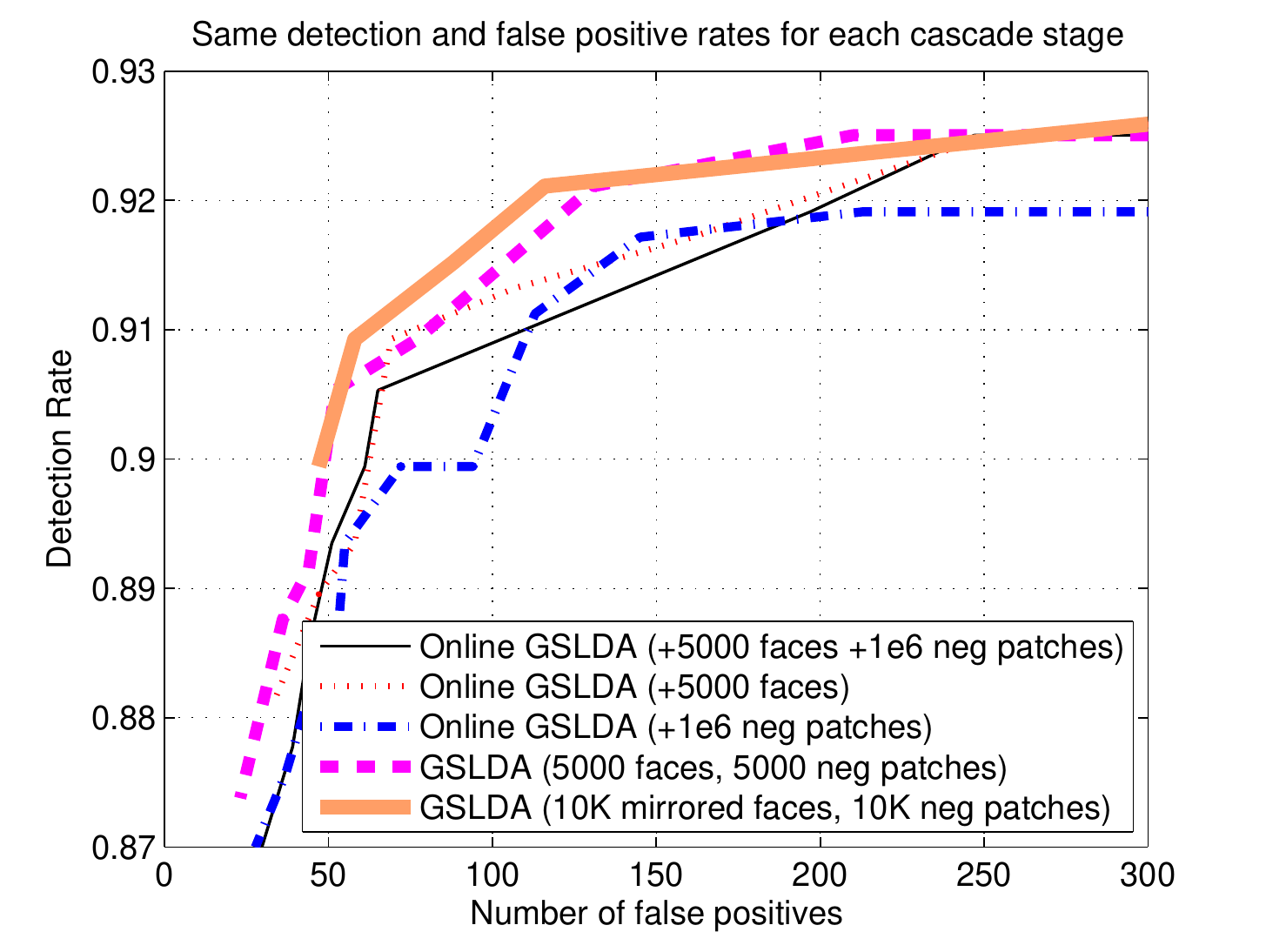}
          \label{fig:cascade_face_c}
        }
        \end{center}
        \caption{
          Comparison of ROC curves on MIT+CMU face test set. The four detectors
          are trained using (a) $500$ faces, (b) $1,000$ faces and (c)
          $5,000$ and $10,000$ mirrored faces.
        }
	    \label{fig:cascade_face}
    \end{figure*}

    \begin{figure}[t!]
      \begin{center}
        \includegraphics[width=0.4\textwidth]{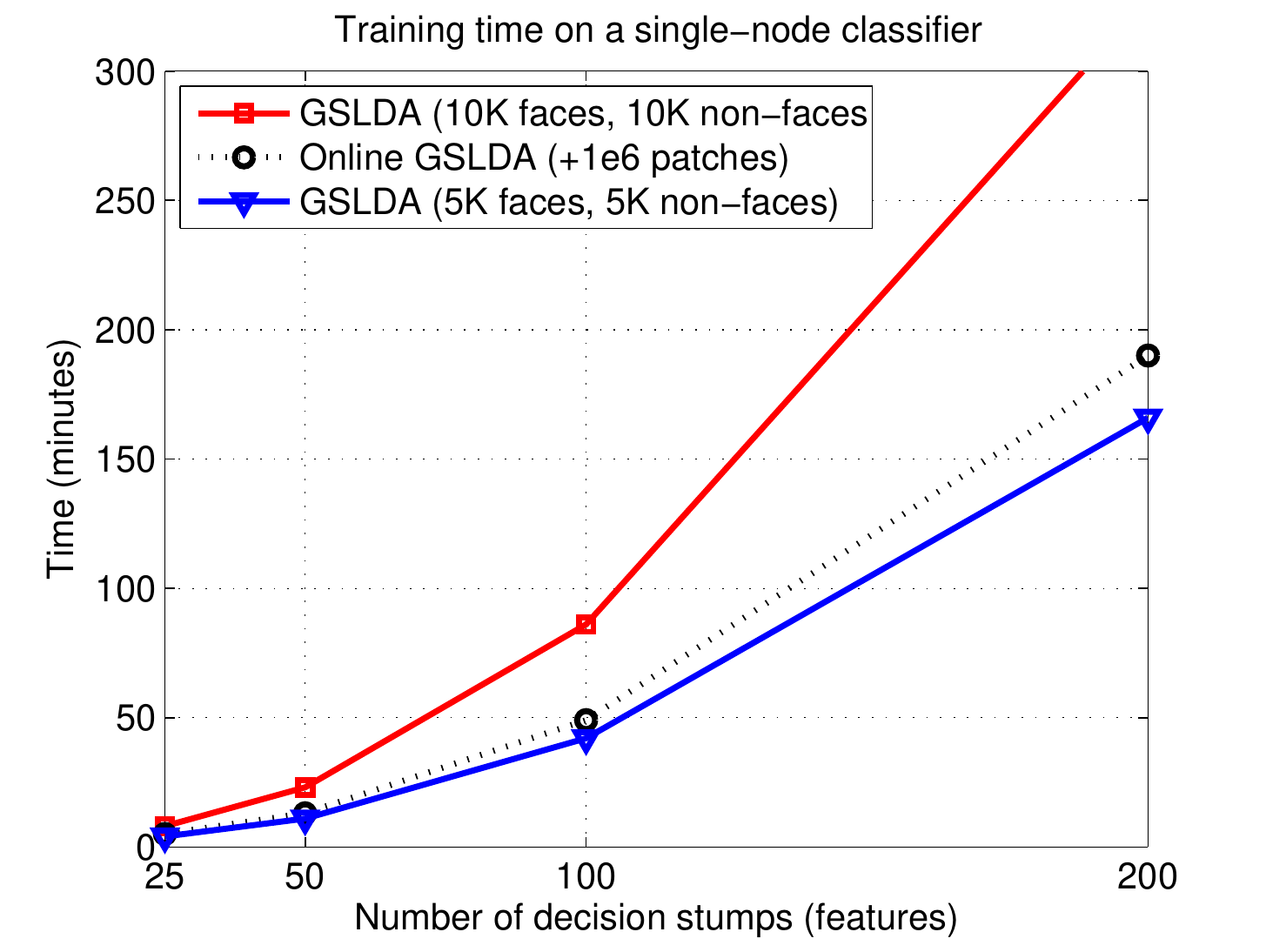}
      \end{center}
         \caption{Comparison of the training time between GSLDA 
         and online GSLDA algorithms. The first and second GSLDA detectors
         are trained with $5,000$ faces and $5,000$ non-faces,
         and $10,000$ faces and $10,000$ non-faces, respectively.
         Online GSLDA is initially trained with $5,000$ faces and
         $5,000$ non-faces and updated with one million new patches.
         Notice that there is a slight increase in training time even though
         we incrementally update with $200 \times$ more training samples.
         }
       \label{fig:gslda_cpu}
     \end{figure}

    In this experiment, we use mirrored faces from previous experiment for batch learning and online
    learning. The number of initial positive samples used in each experiment is varied. We use $500$
    faces, $1,000$ faces and $5,000$ faces to initially train a face detector.  In each experiment,
    we trained four different cascaded detectors. The first cascaded detector is the same as in
    Viola and Jones \cite{Viola2004Robust} \ie, the face data set used in each cascade stage is the
    same while the non-face samples used in each cascade layer are collected from false positives of
    the previous stages of the cascade (bootstrapping). The cascade training algorithm terminates
    when there are not enough negative samples to bootstrap. 

    The second, third and forth face detectors are trained initially with the technique similar to
    the first cascaded detector. However, the second cascaded face detector is incrementally updated
    with new negative examples collected from false positives of the previous stages of cascade. The
    third cascaded face detector is incrementally updated with $5,000$ unseen faces. The final face
    detector is incrementally updated with both false positives from previous stages and unseen
    faces. For each face detector, weak classifiers are added to the cascade until the predefined
    objective is met. In this experiment, we set the minimum detection rate in each cascade stage to
    be $99\%$ and the maximum false positive rate to be $50\%$. 

    We tested our face detectors on the low resolution faces datasets, MIT+CMU frontal face test
    sets. The complete set contains $130$ images with $507$ frontal faces. In this experiment, we
    set the scaling factor to $1.2$ and window shifting step to $1$. The techniques used for merging
    overlapping windows is similar to \cite{Viola2004Robust}. Detections are considered true or
    false positives based on the area of overlap with ground truth bounding boxes. To be considered
    a correct detection, the area of overlap between the predicted bounding box and ground truth
    bounding box must exceed $50\%$. Multiple detections of the same face in an image are considered
    false detections.

    Fig.~\ref{fig:cascade_face} shows a comparison between the ROC curves produced by online GSLDA
    classifier. The ROC curves in Fig.~\ref{fig:cascade_face_a} show that online GSLDA classifier
    outperforms GSLDA classifier at all false positive rates when initially trained with $500$
    faces. Incrementally updating the GSLDA model with unseen faces ($+5000$ faces) yields a better
    result than updating the model with new false positives from previous stages of the cascade
    ($+10^6$ negative patches). The online classifier performs best when updated with both new
    positive and negative patches. Fig.~\ref{fig:cascade_face_b} shows a comparison when the number
    of initial training samples have been increased to $1000$ faces. The performance gap between
    GSLDA and online GSLDA is now smaller. We observe the performance of both GSLDA and online GSLDA
    ($+10^6$ negative patches) to be very similar. This indicates that the cascade learning
    framework proposed by Viola and Jones might have already incorporated the benefit of massive
    negative patches. Incremental learning with new negative instances do not seem to improve the
    performance of cascaded detectors any further. Another way to explain the results of our
    findings is to use the concept of linear asymmetric classifier (LAC) proposed in
    \cite{Wu2008Fast}. In \cite{Wu2008Fast}, the asymmetric node learning goal is expressed as
\begin{align}
  \label{EQ:asymgoal}
  \maximize_{ \bw,w_0}   \quad
        &
        {\Pr}_{\bx \sim (\bm_1,\Sigma_1)} \left\{ \bw^\T \bx \geq w_0 \right\}, \\
        \st \quad &
        {\Pr}_{\by \sim (\bm_2,\Sigma_2)} \left\{ \bw^\T \by \leq w_0 \right\} = \beta. \notag	
\end{align}
    Since, the problem has no closed-form solution, the authors developed an approximate solution when
    $\beta = 0.5$. To find a closed-form solution, the authors assumed that $\bw^\T \bx$ is Gaussian
    for any $\bw$, class $C_2$ distribution is symmetric and the median value of the class $C_2$
    distribution is close to its mean. The direction $\bw$ can then be approximated by 
    \begin{align}
      \label{EQ:LAC}
      \maximize_{\bw \neq 0} \quad \frac{ \bw^\T (\bm_1 - \bm_2)} 
          {\sqrt{ \bw^\T \Sigma_1 \bw}}.
    \end{align}
    From their objective functions, the only difference between FDA \eqref{EQ:FDA} and LAC
    \eqref{EQ:LAC} is that the pooled covariance matrix of FDA, $\Sigma_1$ + $\Sigma_2$, is replaced
    by the covariance matrix of class $C_1$, $\Sigma_1$. In other words, when train the classifier
    with the asymmetric node learning goal for the cascade learning framework, the variance of
    negative classes becomes less relevant. In contrast, new instances of positive classes affect
    both the numerator and denominator in \eqref{EQ:LAC}. Hence, it is easier to notice the
    performance improvement when new positive instances are inserted. Our results seem to be
    consistent with their derivations.

    We further increase the number of initial training faces to $5,000$. All face detectors now seem
    to perform very similar to each other. We conjecture that this is the best performance that our
    cascaded detectors with the provided training set can achieve on MIT+CMU data sets. The results
    of the face detectors trained with $10,000$ faces and $10,000$ non-faces seem to support our
    assumptions (Fig.~\ref{fig:cascade_face_c}). To further improve the performance, different
    cascade algorithms, \eg, soft-cascade \cite{Bourdev05SoftCascade}, WaldBoost
    \cite{Sochman2005WaldBoost}, multi-exit classifiers \cite{Pham2008Detection}, \etc. and a
    combination with other types of features, \eg, edge orientation histograms (EOH)
    \cite{Levi2004Learning}, covariance features \cite{Shen2008Face}, \etc,
    can also be experimented.
    Fig.~\ref{fig:gslda_cpu} shows a comparison of the computation cost between batch GSLDA and
    online GSLDA. The horizontal axis shows the number of weak learners (decision stumps) and the
    vertical axis indicates the training time in minutes. From the figure, online learning is much
    faster than training a batch GSLDA classifier as the number of weak learners grows. On average,
    our online classifier takes less than $1.5$ millisecond to update a strong classifier of $200$
    weak learners on standard off-the-shelf PC with the use of GNU scientific library (GSL)\footnote
    {\url{http://www.gnu.org/software/gsl/} }.

\section{Conclusion}
\label{sec:conclusion}

In this work, we have proposed an efficient online object detection algorithm. Unlike many existing
algorithms which applied boosting approach, our framework makes use of greedy sparse linear
discriminant analysis (GSLDA) based feature selection which aims to maximize the class-separation
criterion. Our experimental results show that our incremental algorithm does not only perform
comparable to batch GSLDA algorithm but is also much more efficient. On USPS digits data sets, our
online algorithm with decision stumps weak learners outperforms online boosting with
class-conditional Gaussian distributions. Our extensive experiments on face detections reveal that
it is always beneficial to incrementally train the detector with online samples. Ongoing works
include the search for more accurate and efficient online weak learners.

\bibliographystyle{ieee}



\begin{IEEEbiography}{Sakrapee Paisitkriangkrai} 
        is currently pursuing Ph.D. degree at the University of New
        South  Wales, Sydney, Australia. He received the B.E. degree
        in computer engineering and M.E. degree in biomedical
        engineering from the University of New South Wales in 2003.
        His research interests include pattern recognition, image
        processing and machine learning. 
\end{IEEEbiography}

\begin{IEEEbiography}{Chunhua Shen}
        received the Ph.D. degree from School of Computer Science,
        University of Adelaide,
        Australia, in 2005.  Since Oct. 2005, he has been
        a researcher with the
        computer vision program, NICTA (National ICT Australia),
        Canberra Research Laboratory.
        He is also an adjunct research fellow at the
        Australian National University; and adjunct lecturer at the
        University of Adelaide. His main research interests include
        statistical machine learning and its applications in computer
        vision and image processing. 
\end{IEEEbiography}

\begin{IEEEbiography}{Jian Zhang} 
        (M'98-SM'04) received the Ph.D. degree in electrical
        engineering from the University College, University of New
        South Wales, Australian Defence Force Academy, Australia, in
        1997.  He is a principal researcher in NICTA, Sydney.  He is
        also a conjoint associate professor at University of New South
        Wales.
        His research interests include image/video
        processing, video surveillance and multimedia content management.
        Dr. Zhang is currently an associate editor of the IEEE
        Transactions on Circuits and Systems for Video Technology and
        the EURASIP Journal on Image and Video Processing.
   \end{IEEEbiography}

\end{document}